%% file: main.tex
\documentclass[10pt,twocolumn,letterpaper]{article}

\usepackage[pagenumbers]{cvpr}

\input{preamble}

\definecolor{cvprblue}{rgb}{0.21,0.49,0.74}
\usepackage[breaklinks,colorlinks,allcolors=cvprblue]{hyperref}

\title{Lost in Reality:\\
Measuring the Form Gap and Distilling Stronger, More Robust VLMs}

\author{%
Yutao Fan\textsuperscript{1,2,*}\quad
Yibin Huang\textsuperscript{1,*}\quad
Heng Zhou\textsuperscript{3}\quad
Songtao Huang\textsuperscript{2}\quad
Li Kang\textsuperscript{4}\\
Jiwen Yu\textsuperscript{5}\quad
Yiran Qin\textsuperscript{6}\quad
Lei Bai\textsuperscript{2}\quad
Wangmeng Zuo\textsuperscript{1}\\[4pt]
{\normalsize
\textsuperscript{1}Harbin Institute of Technology\quad
\textsuperscript{2}Shanghai Artificial Intelligence Laboratory}\\
{\normalsize
\textsuperscript{3}University of Science and Technology of China\quad
\textsuperscript{4}Shanghai Jiao Tong University}\\
{\normalsize
\textsuperscript{5}University of Hong Kong\quad
\textsuperscript{6}The Chinese University of Hong Kong Shenzhen\qquad
{\footnotesize\textsuperscript{*}Equal contribution.}}%
}

\hypersetup{
  pdftitle={Lost in Reality: Measuring the Form Gap and Distilling Stronger, More Robust VLMs},
  pdfauthor={Yutao Fan, Yibin Huang, Heng Zhou, Songtao Huang, Li Kang, Jiwen Yu, Yiran Qin, Lei Bai, Wangmeng Zuo}
}

\makeatletter
\patchcmd{\@maketitle}{\vspace*{24pt}}{\vspace*{12pt}}{}{}
\patchcmd{\@maketitle}{\vspace*{12pt}\end{center}}{\vspace*{2pt}\end{center}}{}{}
\makeatother

\begin{document}
\maketitle
\input{fig/teaser}
\input{sec/0_abstract}
\input{sec/1_intro}
\input{sec/2_related}
\input{sec/3_bench}
\input{sec/3_method}
\input{sec/4_experiments}

\input{sec/6_conclusion}
{
    \small
    \fontsize{9}{10.5}\selectfont
    \setlength{\bibsep}{0.4pt plus 0.4pt minus 0.3pt}
    \bibliographystyle{ieeenat_fullname}
    \bibliography{refs}
}

\end{document}

%% file: preamble.tex
\usepackage{cuted}      %
\usepackage{caption}    %

\usepackage{multirow}
\usepackage{tabularx}
\usepackage{array}
\usepackage[ruled,vlined,linesnumbered]{algorithm2e}

\usepackage{microtype}

\usepackage{colortbl}
\usepackage{pifont}
\newcommand{\cmark}{\ding{51}}
\newcommand{\xmark}{\ding{55}}
\definecolor{oursbg}{RGB}{232,241,251}
\newcommand{\oursrow}{\rowcolor{oursbg}}
\newcolumntype{Y}{>{\centering\arraybackslash}X}
\newcommand{\best}[1]{\textbf{#1}}
\newcommand{\second}[1]{\underline{#1}}

\newcommand{\grp}[2]{\multicolumn{#1}{@{}l}{\textit{\textcolor{gray}{#2}}}}
\newcommand{\hd}[2]{\begin{tabular}[b]{@{}c@{}}#1\\#2\end{tabular}}
\newcommand{\two}{}

\newif\ifshowtodos
\showtodostrue
\ifshowtodos
  \newcommand{\TODO}[1]{\textbf{\color{red}[TODO: #1]}}
\else
  \newcommand{\TODO}[1]{}
\fi

\newcommand{\method}{REVA\xspace}

\newcommand{\Top}{\mathcal{K}}
\DeclareMathOperator*{\argmax}{arg\,max}

\definecolor{cTeacher}{RGB}{230,159,0}
\definecolor{cStudent}{RGB}{0,114,178}
\definecolor{cSame}{RGB}{150,150,150}
\definecolor{cExcess}{RGB}{0,114,178}
\definecolor{cRandom}{RGB}{86,180,233}
\definecolor{cHeld}{RGB}{213,94,0}
\definecolor{cCapture}{RGB}{0,158,115}

\AtBeginDocument{%
  \crefname{algocf}{Alg.}{Algs.}%
  \Crefname{algocf}{Algorithm}{Algorithms}%
}

%% file: fig/teaser.tex
\begin{strip}
  \centering
  \includegraphics[width=\textwidth]{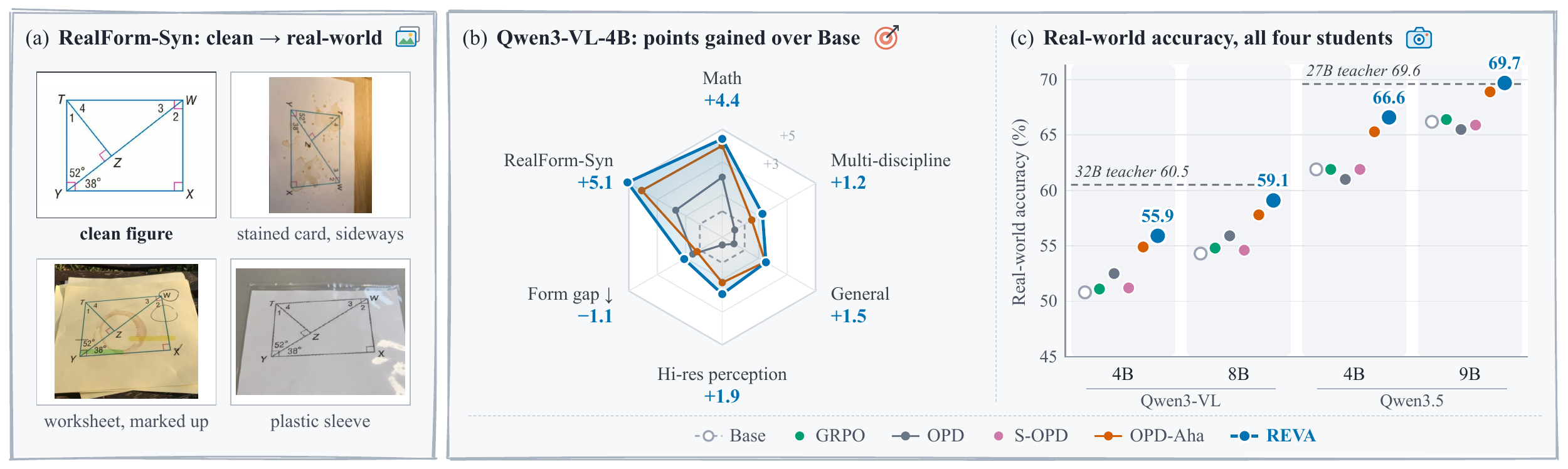}
  \captionsetup{hypcap=false}
  \captionof{figure}{\textbf{RealForm and \method at a glance.} (a) RealForm-Syn composites the exact pixels of a
  benchmark figure into real-world media; question and answer are unchanged. (b) Qwen3-VL-4B: points gained over Base
  by OPD, OPD-Aha and \method on six measures (\cref{tab:main,tab:perception,tab:rfmethods}), on one scale for every
  axis (dashed ring: Base; the form gap drawn outward when it shrinks); numbers are \method's. (c) Real-world accuracy on RealForm-Syn: \method leads
  for every student and is level with the 27B teacher on Qwen3.5-9B (dashed: the teachers).}
  \captionsetup{hypcap=true}
  \label{fig:teaser}
\end{strip}

%% file: sec/0_abstract.tex
\begin{abstract}
Vision-language models (VLMs) have become strong at both perceiving and reasoning about charts, diagrams and math
figures on standard benchmarks, which show each figure as a clean digital image. In the real world, however, the same
content often arrives as a printed page, a screen, a projected slide or a hand-drawn copy, and this change of visual
form can lower their accuracy. Existing benchmarks either corrupt whole images without a medium or cover only one to
four captured forms, often photographing the question text as well. To cleanly isolate this form gap under a fixed
prompt, we build RealForm: RealForm-Syn composites the exact pixels of 1,580 benchmark figures into 4,551 images of 15
real-world media, and RealForm-Real captures a stratified subset of the same figures as real photographs on screens and
paper. All 23 open models we evaluate, from 2B to 397B parameters, lose about 12 points on average on both parts.
Distilling a stronger teacher into a smaller VLM via on-policy distillation (OPD) gives no supervision on other forms
when both read clean images, whereas giving only the student a harder view can supervise it on clean-image details its
view barely shows, teaching ungrounded guesses. To overcome this, we propose \method (Relative-Evidence distillation
with View Asymmetry). For each training image, \method selects a harder view that challenges the student while staying
clear to the teacher. Its target keeps the teacher's clean-image prediction, reweighted by contrasting this view with a
blank image. This discounts language priors and focuses supervision on visual tokens that the view still supports. On
four students from two families, \method improves over OPD by 1.9--2.5 points on the mean of ten standard benchmarks
and by 1.5--3.0 points on six high-resolution perception benchmarks. On both RealForm-Syn and RealForm-Real, it is the
most accurate method for every student, 3.2--5.6 and 3.3--6.3 points above OPD, respectively.
\end{abstract}

%% file: sec/1_intro.tex
\section{Introduction}
\label{sec:intro}

Vision-language models (VLMs) have become strong at both perceiving and reasoning about visual content. Recent open
models~\cite{qwen3vl2025,qwen35,internvl35,glm45v,kimivl} read values off charts, follow the construction in a geometry
diagram and solve competition math problems stated in figures~\cite{lu2024mathvista,wang2024mathvision,zhang2024mathverse},
answer college-level questions across disciplines~\cite{yue2024mmmu}, and find small details in 4K and 8K
images~\cite{wu2024vstar,wang2025hrbench}. Reinforcement learning on problems with checkable answers has further
improved their reasoning~\cite{glm45v,internvl35,vlrethinker2025}.
These abilities have brought VLMs into wide use. Assistants on phones answer questions about what the camera sees,
tutoring apps explain math and science problems, document tools read scanned and photographed
pages~\cite{wilddoc2025,docptbench2025}, and agents read and operate computer and phone
screens~\cite{qwen3vl2025,glm45v,kimivl}.

Most benchmarks, however, show each figure as a clean digital image, while in these uses the figure often reaches
the model in another form: a student photographs a homework problem on a printed page, a listener photographs a
projected slide or a screen, and a teacher copies a diagram by hand onto a whiteboard.
Some benchmarks do include such inputs, with simulated corruptions of the whole image~\cite{mmcbench2024,rbench2024}
or real captures of documents, charts and math problems~\cite{wilddoc2025,mathscape2024,mathreal2025,mathsight2025},
and most find VLMs less accurate on them. We refer to this accuracy drop, measured on the same question from its
clean figure to another form, as the \emph{form gap}.
Existing benchmarks, however, neither span diverse forms nor cleanly isolate this gap (\cref{tab:benchcmp}). Whole-image
corruptions place the figure on no medium, while costly real captures cover only one to four
forms~\cite{wilddoc2025,mathreal2025,mathscape2024,mathsight2025,mathvisionwild}. Moreover, rather than changing only
the figure under a fixed text prompt, several photograph the printed problem statement as
well~\cite{mathscape2024,mathreal2025,mathvisionwild}, confounding figure perception with text OCR. On MathReal and
MathVision-Wild, some models even score higher on the photos than on the clean
originals~\cite{mathreal2025,mathvisionwild}.

To measure the form gap, we build RealForm (\cref{sec:bench}), which changes only the figure.
RealForm-Syn takes 1,580 questions that need the image from six benchmarks and places each figure on three
of 15 real-world media: nine kinds of paper, five screens and projected slides (\cref{fig:rfstats}). A generative
model draws only the empty medium. The original figure is then warped onto it, so no label or number changes. The
figure picks up the medium's lighting and stains, as well as printing and phone-capture
effects such as blur and moir\'e.
Each of the 4,551 images is paired with its clean figure, labeled with its conditions and screened so that the
needed information stays readable. To check these against real captures, RealForm-Real photographs a stratified
sample of 120 of the 1,580 questions (20 per source) on screens and paper.
Across 23 open models from 2B to 397B parameters, each loses 7.6--15.0 points on RealForm-Syn and 12.4 points on
average on RealForm-Real, matching the 11.7-point loss on the composited images of the same questions and media.
The three strongest closed-source models lose only 1.6--4.4 points on RealForm-Syn and 1.2--3.0 on RealForm-Real, so the
gap is not inevitable
(\cref{sec:exp:existing}).

Small VLMs used in practice are often trained by on-policy distillation
(OPD)~\cite{agarwal2024gkd,tm2025opd,visionopd2026,vaopd2026,fpopd2026,opdaha2026}, where a stronger teacher scores
each token of the student's responses.
In standard OPD, both models read the clean image: distilling Qwen3-VL-32B into Qwen3-VL-4B recovers only 1.7 of
the teacher's 9.7-point real-world lead, and the form gap barely moves (\cref{fig:teaser}c, \cref{tab:rfmethods}).
Contrasting clean-image predictions with a blank image~\cite{opdaha2026,vcsd2026} sharpens visual supervision yet
still shows no other forms, whereas feeding harder views to both models exposes the target to the teacher's own
form gap. A privileged teacher~\cite{lopezpaz2016unifying,chen2019lbc,nopd2026,s2vopd2026,scopd2026} lets only the
student read a harder view, but targets or contrast weights taken from the clean image then rest on details the
view barely shows, teaching the student to claim image content even on a blank gray image (\cref{sec:exp:dilemma}).

To resolve this mismatch, we propose \method (Relative-Evidence distillation with View Asymmetry, \cref{fig:method}).
For each training image, \method makes several harder views and picks the one that hurts the student most relative to
the teacher. Views that also hurt the teacher are penalized, keeping inputs that stay clear to the teacher yet
challenge the student.
The target keeps the teacher's clean-image prediction, reweighted by contrasting the \emph{student's view} with a
blank image~\cite{leng2024vcd,favero2024m3id,opdaha2026,vcsd2026}. This discounts language priors and focuses
supervision on visual tokens that the view still supports.
Across four students from two model families, \method
achieves the best results on standard benchmarks, high-resolution perception and both parts of RealForm. Over
OPD-Aha (\method without views), its gain grows from 0.4--0.5 points on standard benchmarks to 1.1 on RealForm-Syn,
with a 1.4-point smaller form gap, and 1.7 on RealForm-Real.

Our contributions are as follows:
\begin{itemize}
  \item \textbf{We introduce RealForm to isolate the form gap across diverse media.} Keeping the text prompt fixed, it
pairs clean figures with composited images across 15 media and with real photos. All 23 open VLMs lose about 12 points
on average on both parts.
  \item \textbf{We propose \method for on-policy distillation under view asymmetry.} It selects harder views that
challenge the student while staying clear to the teacher. Contrasting each view with a blank image then discounts
language priors and weights visual tokens the view still supports.
  \item \textbf{We demonstrate consistent gains across model families, scales and forms.} Across four students, \method
leads all baselines on ten standard benchmarks, high-resolution perception and RealForm. Notably, 4B \method students
match or surpass 8B/9B OPD students while avoiding ungrounded answers on gray images.
\end{itemize}

%% file: sec/2_related.tex
\section{Related work}
\label{sec:related}

\begin{figure*}[t]
  \centering
  \includegraphics[width=\textwidth]{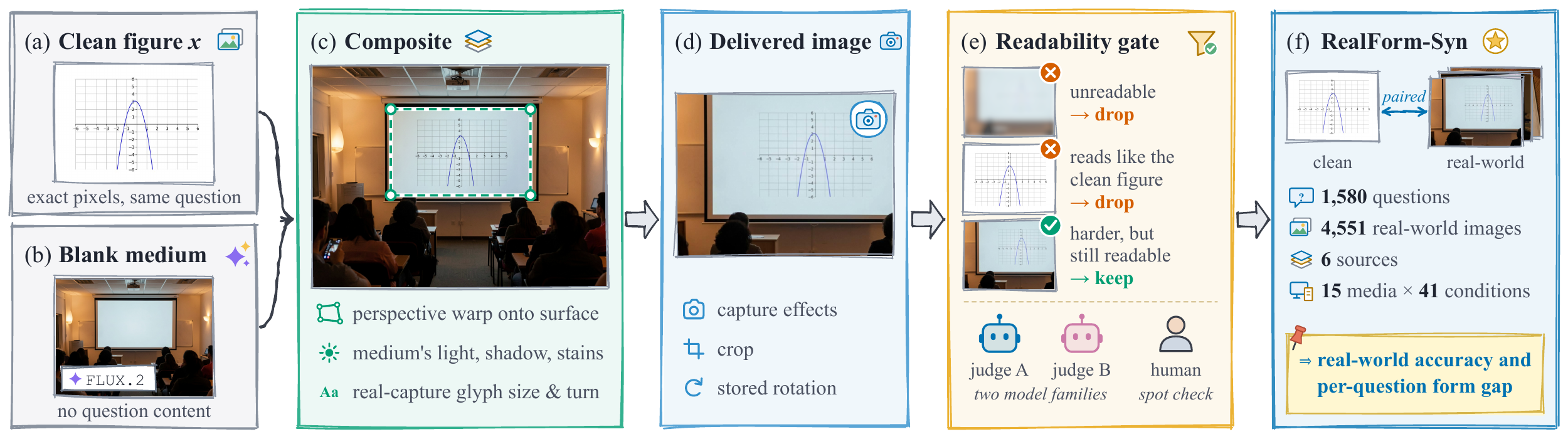}
  \caption{\textbf{Building RealForm-Syn.} Only the empty medium (b) is generated; the clean figure (a) is warped onto
  it with its own pixels (c) and takes on its light and stains, so no label or number changes. (d) Capture effects and
  the stored rotation give the image the model sees. (e) Two judges remove unreadable and too-easy images. (f) 1,580
  questions with 4,551 real-world images on 15 media.}
  \label{fig:rfpipe}
\end{figure*}

\paragraph{On-policy distillation.}
GKD~\cite{agarwal2024gkd}, MiniLLM~\cite{gu2024minillm} and the Thinking Machines recipe~\cite{tm2025opd} train the
student on its own samples under token-level teacher supervision; later work varies the
divergence~\cite{ko2024distillm,opdsurvey2026} or weights tokens~\cite{tip2026}. For VLMs, VA-OPD~\cite{vaopd2026}
up-weights tokens whose teacher probability drops on a pixelated image, with no gain over uniform weights in a later
comparison~\cite{vad2026}; OPD-Aha~\cite{opdaha2026} and VCSD~\cite{vcsd2026} contrast the teacher's
prediction with its prediction on a blank image, as visual contrastive decoding does at
inference~\cite{leng2024vcd,favero2024m3id}; VAD~\cite{vad2026} keeps the part of the teacher's correction that follows
the change from an evidence crop to its degraded copy, FP-OPD~\cite{fpopd2026} projects it onto visual directions the
student can reach, and PCD~\cite{pcd2026} separates perception from reasoning errors; VC-OPD~\cite{vcopd2026} weights tokens by contrasting the teacher on
the original and on a counterfactual image, and OPD-V~\cite{opdv2026} selects them by the margin between teachers on
a zoomed-in and a masked view. All assume that the student sees
the evidence the teacher's prediction rests on; with a harder student view, evidence measured on the teacher's image may
be missing, and amplifying it rewards unsupported answers. \method keeps OPD-Aha's form of the target but measures the
evidence in the student's own view.

\paragraph{Privileged teachers and asymmetric inputs.}
Teachers that see more than their student are classic in distillation~\cite{lopezpaz2016unifying,chen2019lbc,
xie2020noisystudent}, but a student cannot follow a teacher on what it does not observe (the imitation
gap~\cite{weihs2021advisor}); DOPD~\cite{dopd2026} likewise separates the capability gap a student can close from an
information gap it can only mimic. Recent VLM methods give the teacher an evidence crop (Vision-OPD~\cite{visionopd2026},
PVD~\cite{pvd2026}) or the student a corrupted, low-resolution or token-pruned image (NOPD~\cite{nopd2026},
RP-OPSD~\cite{rpopsd2026}, S$^2$VOPD~\cite{s2vopd2026}, LT-OPD~\cite{ltopd2026}, SCOPD~\cite{scopd2026}), mostly with
the model itself as teacher, which teaches invariance rather than a stronger reading; S-OPD~\cite{sopd2026} keeps an
external teacher on the original image and adds student-side terms on masked and noised copies. All keep the teacher's
distribution on its own input as the target, however much evidence the student's view still shows, and evaluate on
clean images or under the training reduction. \method selects views harder for the student than for the teacher and
amplifies only evidence the teacher still finds in them.

\paragraph{Robustness to rendering and capture.}
Robustness of vision models is measured under synthetic corruptions~\cite{hendrycks2019imagenetc}, on renditions
such as sketches~\cite{hendrycks2021imagenetr,wang2019imagenetsketch} and on controlled real
photographs~\cite{barbu2019objectnet}. For multimodal models, MMCBench~\cite{mmcbench2024} and
others~\cite{qiu2024mmshift} apply image corruptions, R-Bench~\cite{rbench2024} a simulated capture chain and
Q-Bench~\cite{wu2024qbench} low-level distortions; similar benchmarks cover documents~\cite{chen2024rodla} and
charts~\cite{moured2025chaos}, while DynaMath~\cite{zou2024dynamath} changes a figure's content, and with it the
answer, by program. Augmentation~\cite{hendrycks2020augmix,hendrycks2021imagenetr} and degradation-aware
reasoning~\cite{robustr12025} train for such shifts. None of these keeps a figure's pixels on a real-world medium, so redrawing,
printing, screens and projection of the same figure fall outside them. Real captures cover these forms, from
print-and-photograph adversarial examples~\cite{kurakin2017physical} to benchmarks of documents and
charts~\cite{wilddoc2025}, photographed documents~\cite{docptbench2025}, math pages~\cite{mathscape2024}, phone photos of K-12 problems~\cite{mathreal2025} and
hand-redrawn figures~\cite{mathsight2025}; they are costly, some lack a paired clean figure and they serve only for
evaluation, while dewarping~\cite{ma2018docunet} flattens a photographed page before it is read. RealForm composites the
original figure pixels into real-world media, so every item is paired with its clean figure and task ability is
separated from the form gap; we also use re-renderings, including editor redraws~\cite{qwenimage2025}, as a
distillation signal.

%% file: sec/3_bench.tex
\section{RealForm: paired form evaluation}
\label{sec:bench}

RealForm has two parts: composited real-world images in RealForm-Syn (\cref{sec:bench:syn}) and real captures in
RealForm-Real (\cref{sec:bench:other}). Both keep each question, its options and its answer from the source
benchmark. Only the figure changes, so every image is compared with its own clean figure. Compared with existing
evaluations of visual form (\cref{tab:benchcmp}), RealForm covers more kinds of form, keeps the question text out of
the image and labels the conditions of every composited image.

\begin{table}[t]
  \centering
  \scriptsize
  \caption{\textbf{RealForm against existing evaluations of visual form.} Synth./Real: the figure on a synthesized or
  captured real-world medium, not a corrupted whole image. Forms: kinds / images per question. Q.\ text: question not
  photographed. Paired: clean figure per item (part: for some). Labels: per-image conditions. --: none/not reported.}
  \label{tab:benchcmp}
  \setlength{\tabcolsep}{2.2pt}
  \input{tables/bench_compare}
\end{table}

\subsection{Questions and the form gap}
\label{sec:bench:items}
We draw 300 questions from each of six perception-heavy benchmarks with checkable answers, MathVista, MathVision,
Geometry3K, MMStar, DynaMath and MathVerse Vision-Only (MV-VO), stratified by question type and source.
Questions that a text-only model answers correctly without the image are replaced by new draws, so every question needs the
figure. A model's \emph{form gap} (the drop in our tables) is its accuracy on the clean figures minus its accuracy on
the real-world images, with each image compared against the clean figure of its own question.

\subsection{RealForm-Syn: composited real-world images}
\label{sec:bench:syn}
\begin{figure}[t]
  \centering
  \includegraphics[width=\linewidth]{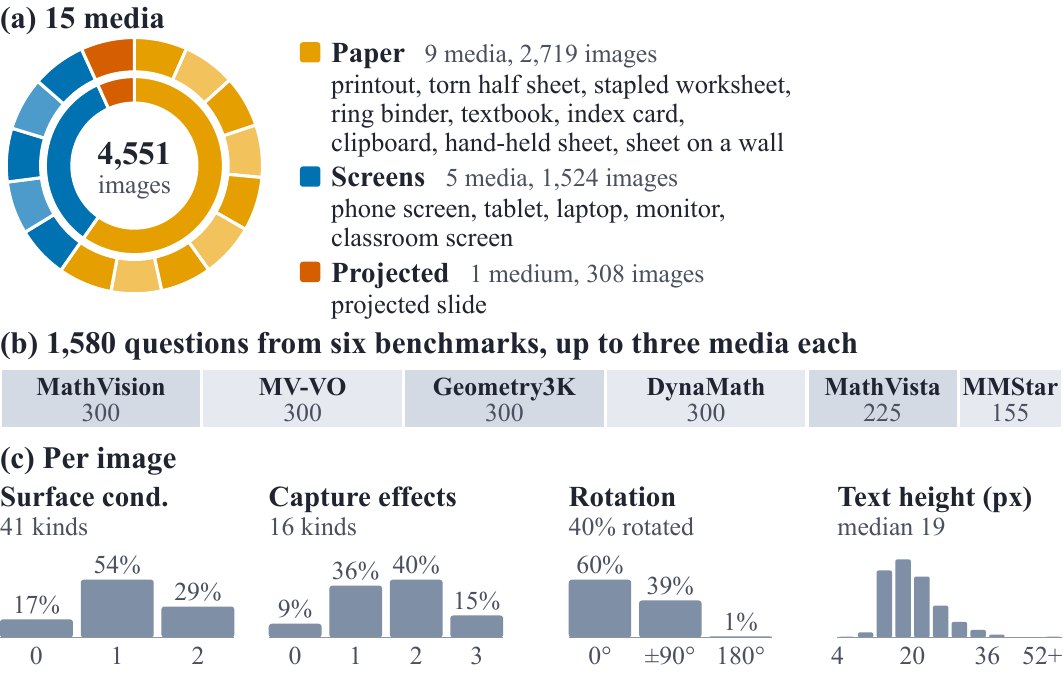}
  \caption{\textbf{Composition of RealForm-Syn.} (a) Images by medium (289--312 each). (b) Questions by source. (c) Per
  image: numbers of surface conditions and capture effects, rotation and text height.}
  \label{fig:rfstats}
\end{figure}
A generative model that redraws a figure can silently change a label or a number, so RealForm-Syn never regenerates
the figure (\cref{fig:rfpipe}). FLUX.2 [klein]~\cite{flux2klein} renders only \emph{empty} media, 15 kinds from
printed sheets to phones, monitors and projected slides (\cref{fig:rfstats}). The original figure is warped and
multiplied into the surface, so the surface's light and stains fall on it while every label and number stays the
original; printing and capture effects are then added, and text heights and turns follow MathReal's user
captures~\cite{mathreal2025}.
Two VLM judges then filter the set. They drop questions whose figure is a natural photograph rather than a chart or
diagram (only in MathVista and MMStar),
real-world images that leave the needed information unreadable, and upright ones as easy to read as the clean figure.
This leaves 1,580 questions and 4,551 real-world images, 96\% readable in a manual sample check. Keeping needed
details readable ensures that drops reflect sensitivity to visual form rather than missing content. Moreover, its media,
surface conditions and 90$^\circ$/270$^\circ$ turns never appear in our training views.

\subsection{RealForm-Real: real captures}
\label{sec:bench:other}
RealForm-Real measures the form gap on real photos and checks whether RealForm-Syn reproduces it. Keeping the prompt
digital, it draws 20 questions per source (120 in all). Each is photographed across three screen types, and 47 also on
folded, crumpled or bent paper, giving 167 captures across viewpoints and lighting.

%% file: tables/bench_compare.tex
\begin{tabular}{@{}ll cc c ccc@{}}
\toprule
Benchmark & Domain & Synth. & Real & Forms & Q.\ text & Paired & Labels \\
\midrule
MMCBench~\cite{mmcbench2024} & natural & \xmark & \xmark & -- & -- & \cmark & \cmark \\
R-Bench~\cite{rbench2024} & natural & \xmark & \xmark & -- & \cmark & \cmark & \cmark \\
\midrule
WildDoc~\cite{wilddoc2025} & docs & \xmark & \cmark & 2\,/\,4 & \cmark & \cmark & \cmark \\
MathScape~\cite{mathscape2024} & K-12 math & \xmark & \cmark & 2\,/\,-- & \xmark & part & \xmark \\
MathReal~\cite{mathreal2025} & K-12 math & \xmark & \cmark & 1\,/\,1 & \xmark & part & \cmark \\
MathSight~\cite{mathsight2025} & univ.\ math & \xmark & \cmark & 2\,/\,2 & \cmark & \cmark & \xmark \\
MathVision-Wild~\cite{mathvisionwild} & comp.\ math & \xmark & \cmark & 4\,/\,1 & \xmark & \cmark & \xmark \\
\midrule
\oursrow RealForm (ours) & math, general & \cmark & \cmark & 15\,/\,3 & \cmark & \cmark & \cmark \\
\bottomrule
\end{tabular}

%% file: sec/3_method.tex
\section{Method}
\label{sec:method}

\input{fig/method}

\label{sec:method:decomp}
\method (\cref{fig:method}) lets the student generate on a harder view while keeping the teacher's target on the
clean original, avoiding teacher errors on harder forms. It also runs the teacher on the student's view to select
that view (\cref{sec:method:asym}) and to reweight the target by the evidence it still shows (\cref{sec:method:target}).

\subsection{Preliminaries}
\label{sec:method:opd}

Let $q$ be a question with original image $x$, $\pi_\theta$ the student (initialized from $\pi_0$) and $p_T$ a frozen
teacher. The student samples $y\sim\pi_\theta(\cdot\mid v,q)$ on an input image $v$ and reads $c^S_t(v)=(v,q,y_{<t})$,
while the teacher reads $c^T_t=(x,q,y_{<t})$. OPD uses $v=x$; \emph{view asymmetry} gives only the student a harder view $v$. %
Let $\Top_t$ be the top-ranked next tokens of the original-image teacher $p_T(\cdot\mid c^T_t)$. The per-token loss is
a forward KL from the target to the student on $\Top_t$:
\begin{equation}
  d_t(v)=\sum_{i\in\Top_t} e^{\tilde\ell_t(i)}
  \big[\tilde\ell_t(i)-\tilde\ell_\theta(i\mid v)\big],
  \label{eq:loss}
\end{equation}
where $\tilde\ell_\theta(i\mid v)=\log\pi_\theta(i\mid c^S_t(v))$ is the student's log-probability on $v$ and
$\tilde\ell_t(i)$ is the log-target, which in OPD is the original-image teacher's $\log p_T(i\mid c^T_t)$.
The loss is averaged over batch response tokens, with gradients only through
$\tilde\ell_\theta$~\cite{agarwal2024gkd}. With $v=x$, OPD shows no harder forms and weights language priors alongside
visual tokens. With $v\neq x$, $v$ may obscure clean details or hurt the teacher, so matching $\log p_T(i\mid c^T_t)$
rewards ungrounded guesses.

\subsection{Excess view selection}
\label{sec:method:asym}

Independent of RealForm's media, each training question has a bank $\mathcal{B}(q)$ of up to nine candidate views of $x$: eight pixel edits (a mixed
noise, downsampling and gamma corruption~\cite{nopd2026}, mild defocus, simulated capture, glare, scan artifacts,
old-print posterization, a mild shear and a 180$^\circ$ rotation) and a redraw in one of five styles (hand-drawn,
whiteboard, blackboard, grid-paper handwriting, tablet ink) by Qwen-Image-Edit~\cite{qwenimage2025}, prompted to keep
every number and label.
For each candidate $v$, the base student $\pi_0$ samples one response $y$ on $v$. With the $k_3$
estimator~\cite{schulman2020klapprox} $D(r)=r-1-\log r$ of the probability ratio $r$ on each sampled token, we measure
the \emph{student gap} $g_S(v)$ and \emph{teacher gap} $g_T(v)$ from the clean-image teacher and select the view with
the largest \emph{excess}:
\begin{align}
  g_S(v)&=\tfrac{1}{|y|}\textstyle\sum_t D\Big(\frac{p_T(y_t\mid c^T_t)}{\pi_0(y_t\mid c^S_t(v))}\Big),\nonumber\\
  g_T(v)&=\tfrac{1}{|y|}\textstyle\sum_t D\Big(\frac{p_T(y_t\mid c^T_t)}{p_T(y_t\mid c^S_t(v))}\Big),\nonumber\\
  v^\star(q)&=\argmax_{v\in\mathcal{B}(q)}\big[g_S(v)-g_T(v)\big].
  \label{eq:excess}
\end{align}
Penalizing $g_T(v)$ discounts views that degrade the teacher's own reading of $x$, keeping views whose visual evidence
remains clear for target reweighting (\cref{sec:method:target}). If
views were selected by the student gap alone, 36.2\% of the questions would get the 180$^\circ$ rotation; with the
excess, each view is chosen for 9--14\% of the questions.

\subsection{Relative-evidence target}
\label{sec:method:target}

Let $r(v)$ be a blank image of $v$'s size, filled with $v$'s mean color. On the tokens in $\Top_t$, we define the
evidence ratio $\rho_t$ and the target $p^\star_t$, which reweights OPD's target $p_T(\cdot\mid x)$ by $\rho_t$ (question
and prefix omitted):
\begin{align}
  \rho_t(i)&=\frac{p_T(i\mid v)}{p_T(i\mid r(v))},\nonumber\\
  p^\star_t(i)&=m_t\,
  \frac{p_T(i\mid x)\,\rho_t(i)^{\beta}}{\sum_{j\in\Top_t}p_T(j\mid x)\,\rho_t(j)^{\beta}},
  \quad i\in\Top_t,
  \label{eq:evidence}
\end{align}
Here $m_t=\sum_{j\in\Top_t}p_T(j\mid x)$ keeps the total probability on $\Top_t$ unchanged, $\beta$ sets how
strongly the evidence reweights the target, and the log-target in \cref{eq:loss} is $\tilde\ell_t(i)=\log p^\star_t(i)$.
A larger $\rho_t(i)$ means the teacher finds token $i$ more likely on $v$ than on the blank image. %
$\rho$ is the teacher's own estimate, so it also reflects the teacher's language prior and calibration. Under graded
blur, downsampling and occlusion, it falls with each level and follows the teacher's accuracy across conditions.

Contrasting $v$ with $r(v)$ discounts language-prior tokens ($\rho_t(i)\approx1$); normalized over $\Top_t$, visually
grounded tokens supported by $v$ ($\rho_t(i)>1$) gain weight while details lost in $v$ ($\rho_t(i)\le1$) lose share
($\beta=0$ gives OPD's target).
Measuring $\rho$ on $x$ instead of $v$ gives weight to details from $x$ even where $v$ no longer shows them
(\cref{sec:exp:dilemma}). Each training step samples one response per question on
its view and runs the teacher on $(x,q,y)$, $(v,q,y)$ and $(r(v),q,y)$. This adds two teacher forward passes per
response and no generation; views and excess scores are computed once before training.

%% file: fig/method.tex
\begin{figure*}[t]
  \centering
  \includegraphics[width=0.95\textwidth]{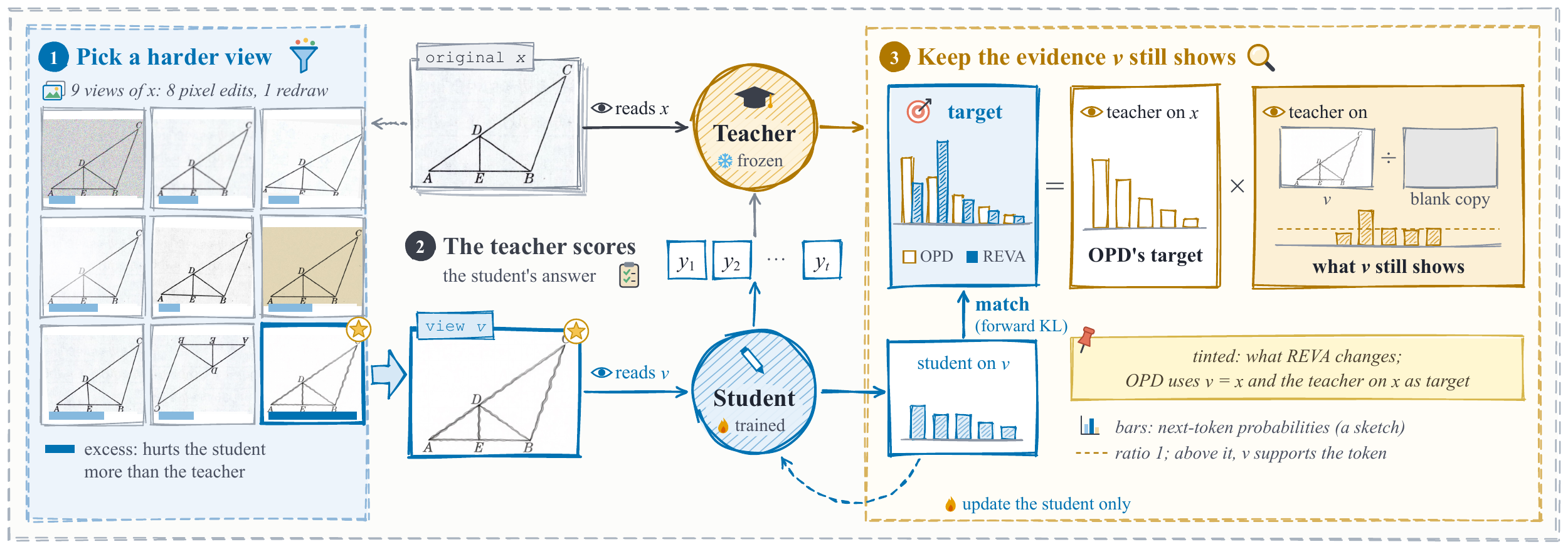}
  \caption{\textbf{Overview of \method.} The teacher's predictions on the student's view are used twice: to pick that
  view and to reweight the target. (1) Before training, the view with the largest excess (\cref{eq:excess}) becomes the student's
  view $v$. (2) The student answers on $v$; the frozen teacher scores the same tokens on the original $x$, giving
  OPD's target. (3) The teacher's ratio $\rho$ between $v$ and its blank copy $r(v)$ (\cref{eq:evidence}) reweights
  OPD's target into \method's target, which the student matches with the forward KL of \cref{eq:loss}.}
  \label{fig:method}
\end{figure*}

%% file: sec/4_experiments.tex
\section{Experiments}
\label{sec:exp}

\subsection{Setup}
\label{sec:exp:setup}

\paragraph{Training.}
The students are Qwen3-VL-4B and 8B, taught by a frozen Qwen3-VL-32B~\cite{qwen3vl2025}, and Qwen3.5-4B and
9B~\cite{qwen35}, taught by a frozen Qwen3.5-27B. Training uses 16,384 single-image ViRL39K
questions~\cite{vlrethinker2025}. \method uses $\beta=2$, and each student's views are selected by its own excess.
We train for 256 steps with 64 questions per step using AdamW and a learning rate of $2\times10^{-6}$,
with 3\% warmup and cosine decay to $10^{-6}$.

\paragraph{Compared methods.}
In \emph{OPD}~\cite{agarwal2024gkd,tm2025opd}, the student and the teacher both read the original image.
\emph{OPD-Aha}~\cite{opdaha2026} uses the target of \cref{eq:evidence} with $v=x$ and \method's $\beta=2$.
\emph{S-OPD}~\cite{sopd2026} adds masking and noise regularizers on the student side. \emph{GRPO} rewards correct
answers and uses no teacher~\cite{vlrethinker2025,shao2024deepseekmath}. For the two 4B students, we also train SFT on the
teacher's correct responses~\cite{kim2016seqkd} and GKD~\cite{agarwal2024gkd}. NOPD~\cite{nopd2026} and
VA-OPD~\cite{vaopd2026} are trained for Qwen3-VL-4B only. We train every method ourselves with the same data, schedule and teacher; NOPD keeps its
own frozen base model as teacher.

\paragraph{Evaluation.}
The standard benchmarks are ten reasoning and multimodal benchmarks: MathVista~\cite{lu2024mathvista},
MathVision~\cite{wang2024mathvision}, WeMath~\cite{qiao2024wemath}, DynaMath~\cite{zou2024dynamath},
LogicVista~\cite{xiao2024logicvista}, Geometry3K~\cite{lu2021intergps}, MMMU~\cite{yue2024mmmu},
MMMU-Pro~\cite{yue2024mmmupro}, MMStar~\cite{chen2024mmstar} and BLINK~\cite{fu2024blink}; Avg.\ is their mean.
High-resolution perception is measured on six benchmarks, V$^*$~\cite{wu2024vstar}, ZoomBench~\cite{wei2026zoombench},
HR-Bench 4K/8K~\cite{wang2025hrbench} and MME-RealWorld EN/CN~\cite{zhang2025mmerealworld}, with the evaluation code
of Vision-OPD and OPD-Aha~\cite{visionopd2026,opdaha2026}; Avg6 is their mean. MathReal~\cite{mathreal2025} gives 175 K-12
problems, each with a clean image and a user's phone capture; we use its official question and judge prompts, with one
sample per item. Decoding follows the
Qwen3-VL model card for vision tasks, with at most 16,384 output tokens. Qwen2.5-72B-Instruct~\cite{qwen25report}
judges the answers of every model on the ten benchmarks, MathReal and RealForm.

\paragraph{RealForm and the gray-image test.}
On RealForm-Syn, each clean figure is answered eight times and each real-world image twice; each capture of
RealForm-Real is answered twice. The gray-image test replaces each of the 800 HR-Bench 4K images with a gray image of
the same size, adds an option to abstain and uses greedy decoding. We report the gray-image answer rate, the percentage
of gray images on which the model neither abstains nor says the object is absent. Lower is better.

\subsection{The form gap of current VLMs}
\label{sec:exp:existing}
\label{sec:bench:eval}

\begin{figure}[t]
  \centering
  \input{fig/rfbars}
  \vspace{-8pt}
  \caption{\textbf{Form gap of 27 VLMs on RealForm.} Each bar gives the accuracy on the clean figures and
  the real-world images of RealForm-Syn; the number above it is the drop, and models are sorted by it. The distance
  from the top of each bar to its orange tick is the model's drop on the real captures of RealForm-Real; a tick
  above the bar marks a gain.}
  \label{fig:rfrefmodels}
\end{figure}
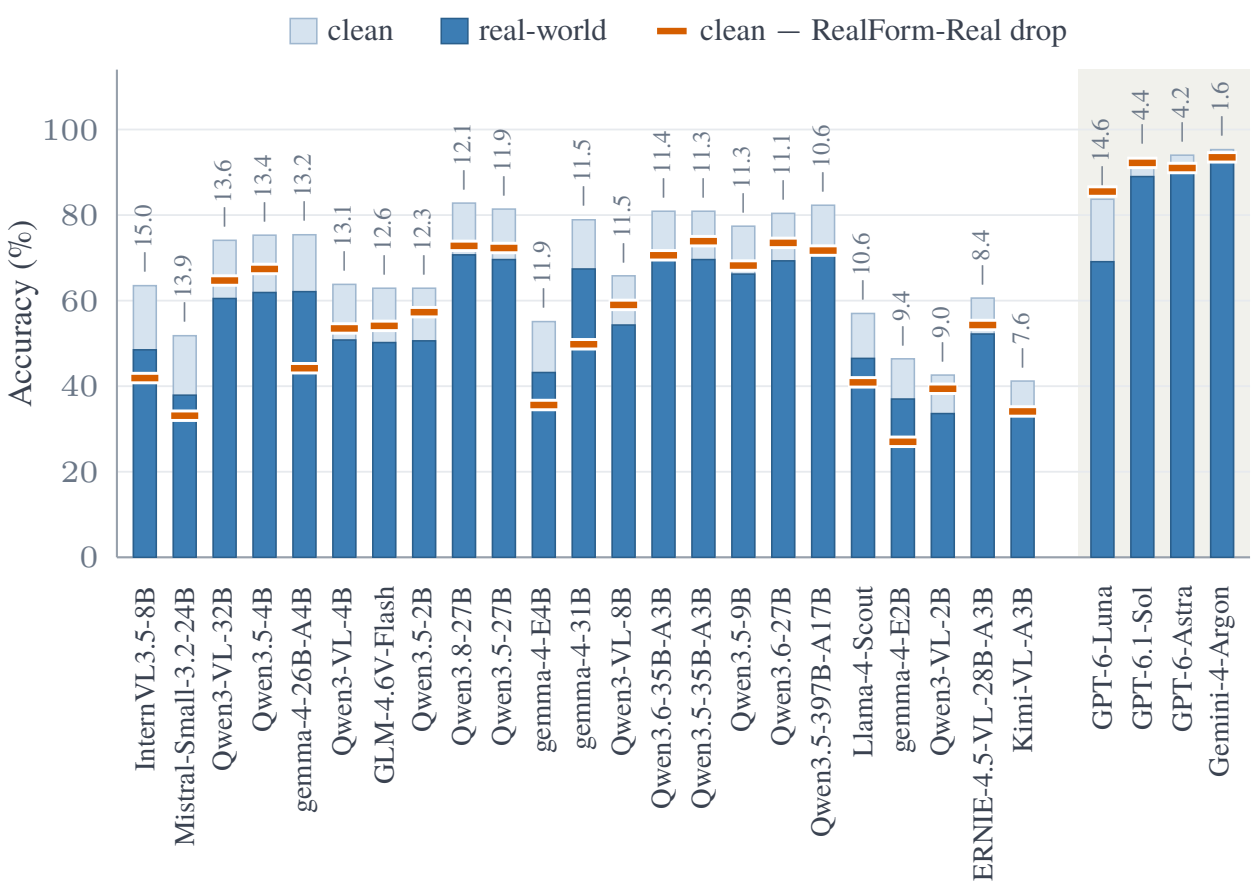

\paragraph{Every open model loses in real-world forms.}
On RealForm-Syn, we evaluate 23 open models from eight families~\cite{qwen3vl2025,qwen35,gemma4,internvl35,glm45v,kimivl,ernie45,mistralsmall32,llama4},
from 2B to 397B parameters, and four closed-source models. Every open model loses 7.6--15.0 points
(\cref{fig:rfrefmodels}). Even the largest open model (Qwen3.5-397B-A17B) loses 10.6 points and the newest
(Qwen3.8-27B) loses 12.1. Our 32B and 27B teachers lose 13.6 and 11.9 points, as much as their 4B--9B students
(11.3--13.4). The three strongest closed-source models lose only 1.6--4.4 points, so the gap is not inevitable;
GPT-6-Luna, the weakest on clean figures, loses 14.6. On RealForm-Real,
open models lose 12.4 points on average, matching their 11.7-point loss on the corresponding
RealForm-Syn images, so RealForm-Syn reproduces this gap.

\paragraph{Turned images and small text cost the most.}
Qwen3-VL-4B loses 21.8 points on real-world images stored sideways or
upside down, against 7.3 on upright ones. It loses 19.1 points on text under 15 pixels high, against 8.9 on text 25
pixels or taller. The loss also grows with the number of capture effects, from 8.3 to 15.0 points. Among the media,
it is largest on book pages and smallest on projections. gemma-4-E4B shows the same pattern, except that paper costs it
about as much as book pages (12.6 and 12.3 points). Rotation is not the whole gap: every open model also loses 4.6--9.1 points on upright images
(11.9--24.1 on turned ones).

\subsection{Main results}
\label{sec:exp:main}
\label{sec:exp:students}

\begin{table*}[t]
  \centering
  \caption{\textbf{Standard benchmarks}: accuracy (\%) on the full benchmarks, one evaluation run; Avg.: mean of the
  ten. Bold / underline: best / second-best per student.}
  \label{tab:main}
  \scriptsize
  \setlength{\tabcolsep}{3.2pt}
  \renewcommand{\arraystretch}{0.9}
  \input{tables/main_results}
\end{table*}

\paragraph{\method is the most accurate on standard benchmarks and perception for every student.}
On the ten benchmarks, \method beats OPD by 1.9--2.5 points (38 of 40 pairs, up to 5.4 on LogicVista) and OPD-Aha by
0.4--0.5 (\cref{tab:main}). All other baselines trail by at least 1.9 points. Against teachers at 74.2 (Qwen3-VL-32B)
and 78.8 (Qwen3.5-27B), \method recovers 36--75\% of the teacher lead over Base, roughly tripling OPD (11--23\%). On
high-resolution perception (\cref{tab:perception}), OPD fails to improve three base models (66.8 \vs 67.9 on
Qwen3-VL-4B). \method gains 1.5--3.0 points over OPD and 0.5--1.1 over OPD-Aha, reaching 75.9 \vs its 27B teacher's
76.1 on Qwen3.5-9B. Notably, 4B \method students match 8B/9B OPD students on Avg.\ (68.6 \vs 68.8; 75.6 \vs 75.5) and
surpass them on Avg6 (69.8 \vs 67.9; 74.6 \vs 73.7).

\begin{table}[t]
  \centering
  \scriptsize
  \caption{\textbf{High-resolution perception}: Avg6, the mean of V$^*$, ZoomBench, HR-Bench 4K/8K and
MME-RealWorld EN/CN. Bold / underline: best / second-best per student (teacher excluded).}
  \label{tab:perception}
  \setlength{\tabcolsep}{5pt}
  \input{tables/perception_avg6}
\end{table}

\label{sec:exp:robust}

\begin{table*}[t]
  \centering
  \caption{\textbf{Robustness to visual form by training method.} RealForm-Syn:
1,580 questions, 4,551 composited real-world images, two samples each; drop: the form gap, clean minus real-world accuracy over the same images (each image against its question's clean figure, eight samples). RealForm-Real: real captures of 120 of these questions, screen and paper together, two samples per capture.
Students: VL, Qwen3-VL; 3.5, Qwen3.5. Bold / underline: best /
second (teacher excluded).}
  \label{tab:rfmethods}
  \footnotesize
  \setlength{\tabcolsep}{4pt}
  \input{tables/rf_methods_wide}
\end{table*}

\paragraph{\method is the most accurate on RealForm-Syn and has the smallest gap.}
Clean-only OPD falls below Base on both Qwen3.5 students, and S-OPD stays within 0.4 points of Base
(\cref{tab:rfmethods}). OPD-Aha lifts accuracy via cleaner perception yet leaves a wider form gap than OPD on
Qwen3-VL (12.9 and 12.3 \vs 12.6 and 10.6). \method raises real-world accuracy by 3.2--5.6 points over OPD and recovers 53--103\% of the
teacher's real-world lead, matching the 27B teacher on Qwen3.5-9B (69.7 \vs 69.6). Averaged over the four students, it
beats OPD-Aha by 1.1 points with a 1.4-point smaller gap. On turned images, whose $90^\circ/270^\circ$ turns never
appear in training, its lead over OPD-Aha doubles to 1.6 points (4.9 over OPD, \vs 0.8 and 3.6 upright).

\paragraph{\method also leads on real captures.}
These gains transfer to real photographs (\cref{tab:rfmethods}). On RealForm-Real, OPD averages only 1.0 point above
Base (1.5 below on Qwen3.5-4B). \method leads on every student, 4.8 points above OPD and 1.7 above OPD-Aha on average,
and equals its 27B teacher on Qwen3.5-9B (72.2). On MathReal phone captures, Base loses 12.3 (Qwen3-VL-4B) and 14.2
points (Qwen3.5-4B). OPD and OPD-Aha leave a 7.8--8.9-point drop, whereas \method cuts it to 4.1 and 5.1 with the
highest capture accuracy on both.

\subsection{Analysis and ablations}
\label{sec:exp:why}
\label{sec:exp:mech}

\begin{figure}[t]
  \centering
  \includegraphics[width=\linewidth]{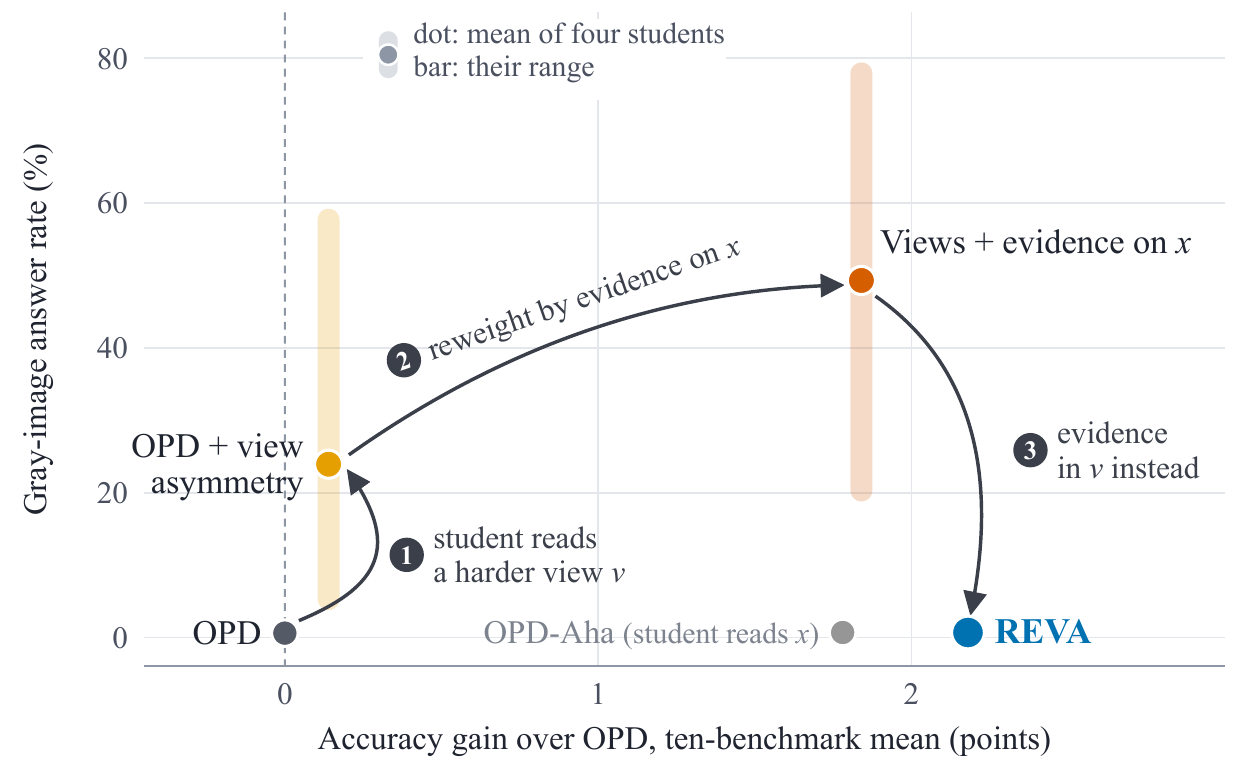}
  \caption{\textbf{Accuracy against the gray-image answer rate} (HR-Bench 4K, greedy decoding). (1) Reading a
harder view raises the gray-image answer rate; (2) reweighting the target by evidence measured on the original $x$ raises
accuracy and this rate; (3) measuring it in the view $v$ (\method) keeps the accuracy while the rate returns
to OPD's level.}
  \label{fig:dilemmamap}
\end{figure}

Reweighting the target by visual evidence gives most of the gain over OPD on the ten benchmarks. OPD-Aha, which
reweights on $x$ without views, already adds 1.5--2.1 points (\method: 1.9--2.5), whereas harder views narrow the form
gap (\cref{tab:rfmethods,tab:ablate}). To see why harder views need view-level evidence, we add \method's steps to OPD
across all four students (\cref{fig:dilemmamap}) and ablate each part on Qwen3-VL-4B (\cref{tab:ablate}).

\paragraph{Reading a harder view teaches the student to answer without evidence.}
\label{sec:exp:dilemma}
\label{sec:exp:components}
Training on \method's views with OPD's target (OPD + view asymmetry) keeps Avg.\ within 0.4 points of OPD. However,
the gray-image answer rate rises from at most 1.1\% under OPD to 3.8\% and 8.8\% on Qwen3-VL and 23.9\% and 59.2\% on
Qwen3.5. Reweighting by clean-image evidence (Views + evidence on $x$) lifts Avg.\ by 1.7--2.0 points over OPD. Yet it
drives the gray-image answer rate to 18.8--79.4\%, because clean-image evidence includes details that $v$ no longer
shows (\cref{fig:dilemmamap}). Within both families, the larger student guesses far more often.

\paragraph{Measuring the evidence in the view removes this cost.}
Computing $\rho_t$ in $v$ instead of $x$ (\method) favors only visual tokens that $v$ still supports. With views
and $\beta=2$ fixed, Avg.\ moves by $-0.1$ to $+0.7$ points against Views + evidence on $x$. The gray-image
answer rate falls to at most 0.9\% for every student, matching OPD and OPD-Aha (at most 1.1 and 1.4\%;
\cref{fig:dilemmamap}). An excess-selected view is thus safe to learn from when evidence is measured where the
student reads.

\begin{table}[t]
  \centering
  \footnotesize
  \caption{\textbf{Main ablation} (Qwen3-VL-4B; columns as in \cref{tab:main,tab:rfmethods}; Clean: accuracy
  on the clean figures over the same images, so Drop = Clean $-$ Real-world). Shaded: at least 1 point worse than
  \method (0.5 for the drop, 3 for the gray-image answer rate). The random-view row keeps \method's view bank and
  target.}
  \label{tab:ablate}
  \setlength{\tabcolsep}{2.8pt}
  \input{tables/ablation_main}
\end{table}

\paragraph{Each part of \method helps.}
Removing the evidence target drops Clean by 3.4 points, Real-world by 3.2 and Avg.\ by 2.1, while raising the
gray-image answer rate to 3.8\% (\cref{tab:ablate}). Removing views leaves Clean unchanged (67.9) but lowers Real-world
by 1.0 point, widening Drop from 12.0 to 12.9. Thus the target lifts overall accuracy while views narrow the form gap.
Both excess selection and the evidence target keep the gray-image answer rate low. With OPD's target, student-gap
selection raises this rate from 3.8\% to 28.8\%. Inside \method, excess beats random views by 0.6--0.7 points on Avg.\
and Real-world and cuts this rate from 2.3\% to 0.9\%.

%% file: fig/rfbars.tex
\definecolor{cRfClean}{HTML}{D6E3EF}\definecolor{cRfCleanE}{HTML}{9FB7CF}\definecolor{cRfReal}{HTML}{3C7DB4}\definecolor{cRfRealE}{HTML}{2A5F8C}\definecolor{cRfTick}{HTML}{D55E00}\definecolor{cRfBand}{HTML}{F1F1EC}\definecolor{cRfInk}{HTML}{3A4250}\definecolor{cRfSub}{HTML}{6B7787}\definecolor{cRfGrid}{HTML}{E7EAEE}
\includegraphics[page=1,width=\linewidth]{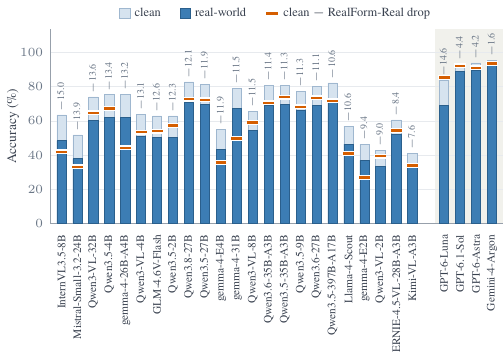}

%% file: tables/main_results.tex
\begin{tabularx}{\textwidth}{@{}c@{\hspace{4pt}}l *{11}{Y}@{}}
\toprule
 &  & \multicolumn{6}{c}{Math} & \multicolumn{2}{c}{Multi-disc.} & \multicolumn{2}{c}{General} & \\
\cmidrule(lr){3-8}\cmidrule(lr){9-10}\cmidrule(lr){11-12}
 & Model & \hd{Math}{Vista} & \hd{Math}{Vision} & \hd{We}{Math} & \hd{Dyna}{Math} & \hd{Logic}{Vista} & \hd{Geo}{3K} & \hd{}{MMMU} & \hd{MMMU}{Pro} & \hd{MM}{Star} & \hd{}{BLINK} & \hd{}{Avg.} \\
\midrule
 & Base & 72.9 & 48.7 & 75.5 & 65.2 & 60.2 & 77.9 & 65.6 & 51.7 & 69.1 & 67.4 & 65.4 \\
 & SFT~\cite{kim2016seqkd} & 75.7 & 45.4 & 75.9 & 65.4 & 54.8 & 80.5 & 63.3 & 49.2 & 67.9 & 63.8 & 64.2 \\
 & GRPO~\cite{vlrethinker2025} & 74.1 & 48.7 & 74.0 & 65.3 & 58.6 & 81.7 & 65.7 & \second{54.1} & 68.7 & 67.5 & 65.8 \\
 & OPD~\cite{agarwal2024gkd,tm2025opd}\two & 75.5 & 50.5 & 78.7 & 67.0 & 58.8 & \second{82.4} & 63.1 & 52.8 & 69.7 & 65.3 & 66.4 \\
 & GKD~\cite{agarwal2024gkd} & 74.1 & 49.9 & 76.5 & 66.5 & 60.4 & 80.7 & 64.8 & 52.5 & 68.9 & 65.8 & 66.0 \\
 & NOPD~\cite{nopd2026} & 73.6 & 49.9 & 76.8 & 65.4 & 56.6 & 82.0 & 64.4 & 52.3 & 69.0 & 67.1 & 65.7 \\
 & VA-OPD~\cite{vaopd2026} & \second{76.4} & \best{51.3} & 78.0 & 66.0 & 58.4 & 81.9 & \second{66.8} & 52.7 & 69.5 & 64.3 & 66.5 \\
 & S-OPD~\cite{sopd2026}\two & 74.4 & 50.1 & 78.7 & 66.7 & \second{61.7} & 79.9 & \best{67.0} & 52.3 & 69.9 & 66.5 & 66.7 \\
 & OPD-Aha~\cite{opdaha2026} & 76.2 & 50.7 & \best{80.8} & \second{68.9} & \best{62.4} & \best{85.4} & 65.2 & 53.1 & \second{70.7} & \best{68.7} & \second{68.2} \\
\oursrow \cellcolor{white}\multirow{-10}{*}{\rotatebox[origin=c]{90}{\shortstack{Qwen3-VL\\4B}}} & \method\two & \best{77.4} & \second{50.8} & \second{80.7} & \best{70.2} & \best{62.4} & \best{85.4} & 65.6 & \best{54.2} & \best{71.3} & \second{68.2} & \best{68.6} \\
\midrule
 & Base & 76.8 & 51.7 & 77.8 & 68.0 & 59.3 & 81.0 & 63.9 & 57.9 & 71.1 & 69.5 & 67.7 \\
 & GRPO & 78.0 & 53.3 & 79.0 & 68.0 & 59.3 & 82.4 & 66.1 & 57.6 & 71.5 & 69.6 & 68.5 \\
 & OPD & 77.5 & 52.7 & 79.1 & 67.9 & \second{60.2} & 83.5 & \best{68.3} & \best{58.3} & 71.3 & 69.4 & 68.8 \\
 & OPD-Aha & \best{79.7} & \second{53.6} & \second{80.8} & \second{70.5} & \best{62.0} & \best{87.0} & \second{67.9} & 57.9 & \second{73.2} & \second{70.9} & \second{70.3} \\
\oursrow \cellcolor{white}\multirow{-5}{*}{\rotatebox[origin=c]{90}{\shortstack{Qwen3-VL\\8B}}} & \method & \second{79.5} & \best{53.9} & \best{83.3} & \best{70.9} & \best{62.0} & \second{85.4} & 67.8 & \second{58.2} & \best{73.8} & \best{72.2} & \best{70.7} \\
\midrule
 & Base & 80.7 & 60.6 & 85.2 & 72.0 & 64.0 & 90.5 & 70.3 & 59.6 & 72.7 & 64.0 & 72.0 \\
 & SFT & 79.5 & 56.6 & 83.9 & 75.9 & 64.9 & 86.9 & 67.7 & 58.3 & 70.5 & 57.5 & 70.2 \\
 & GRPO & \second{81.7} & 61.5 & 84.8 & 75.3 & 67.6 & 88.7 & 71.7 & 61.8 & \second{73.9} & 65.7 & 73.3 \\
 & OPD\two & 80.6 & 60.1 & 86.3 & 78.4 & 64.0 & 91.0 & \second{72.4} & \second{62.3} & 72.2 & 64.1 & 73.1 \\
 & GKD & 80.1 & 61.9 & 86.6 & \second{78.6} & 66.0 & \second{91.5} & 69.9 & 60.8 & 73.3 & 63.8 & 73.2 \\
 & OPD-Aha & \best{82.9} & \second{62.8} & \second{88.4} & \best{80.0} & \second{69.1} & \best{93.3} & 72.0 & 62.0 & 73.8 & \best{67.4} & \second{75.2} \\
\oursrow \cellcolor{white}\multirow{-7}{*}{\rotatebox[origin=c]{90}{\shortstack{Qwen3.5\\4B}}} & \method\two & 81.6 & \best{63.5} & \best{88.8} & \best{80.0} & \best{69.4} & \best{93.3} & \best{73.0} & \best{64.2} & \best{75.3} & \second{66.7} & \best{75.6} \\
\midrule
 & Base & 82.6 & 65.5 & 87.4 & 75.0 & 67.1 & 91.8 & 73.2 & 62.1 & 73.3 & 67.3 & 74.5 \\
 & GRPO & 82.1 & 64.3 & 87.8 & 78.2 & \second{70.5} & \second{93.2} & 73.6 & 63.5 & 75.5 & 67.2 & 75.6 \\
 & OPD\two & 81.9 & 64.2 & 88.4 & 79.6 & 68.5 & 91.3 & \second{73.7} & 64.6 & 74.9 & 67.6 & 75.5 \\
 & OPD-Aha & \second{83.0} & \best{67.9} & \best{89.8} & \best{82.0} & \second{70.5} & 92.2 & 73.0 & \best{66.9} & \best{76.5} & \best{70.4} & \second{77.2} \\
\oursrow \cellcolor{white}\multirow{-5}{*}{\rotatebox[origin=c]{90}{\shortstack{Qwen3.5\\9B}}} & \method\two & \best{84.1} & \second{67.1} & \second{89.6} & \second{81.9} & \best{73.2} & \best{94.0} & \best{75.1} & \second{66.6} & \second{75.8} & \second{69.4} & \best{77.7} \\
\bottomrule
\end{tabularx}

%% file: tables/perception_avg6.tex
\begin{tabular}{@{}l cccc@{}}
\toprule
Avg6 & \hd{Qwen3-VL}{4B} & \hd{Qwen3-VL}{8B} & \hd{Qwen3.5}{4B} & \hd{Qwen3.5}{9B} \\
\midrule
\textcolor{gray}{Teacher} & \textcolor{gray}{72.5} & \textcolor{gray}{72.5} & \textcolor{gray}{76.1} & \textcolor{gray}{76.1} \\
\midrule
Base & 67.9 & 68.2 & 70.4 & 73.7 \\
GRPO & 68.8 & 68.8 & 73.8 & \second{75.1} \\
OPD & 66.8 & 67.9 & 73.1 & 73.7 \\
S-OPD & 67.6 & 67.9 & 73.3 & 74.6 \\
OPD-Aha & \second{69.1} & \second{69.7} & \second{74.1} & 74.8 \\
\oursrow \method & \best{69.8} & \best{70.2} & \best{74.6} & \best{75.9} \\
\bottomrule
\end{tabular}

%% file: tables/rf_methods_wide.tex
\begin{tabular}{@{}l cccc cccc cccc@{}}
\toprule
 & \multicolumn{4}{c}{\hd{RealForm-Syn}{real-world accuracy (\%) $\uparrow$}} & \multicolumn{4}{c}{\hd{RealForm-Syn}{drop (points) $\downarrow$}} & \multicolumn{4}{c}{\hd{RealForm-Real}{capture accuracy (\%) $\uparrow$}} \\
\cmidrule(lr){2-5}\cmidrule(lr){6-9}\cmidrule(l){10-13}
Model & \hd{VL}{4B} & \hd{VL}{8B} & \hd{3.5}{4B} & \hd{3.5}{9B} & \hd{VL}{4B} & \hd{VL}{8B} & \hd{3.5}{4B} & \hd{3.5}{9B} & \hd{VL}{4B} & \hd{VL}{8B} & \hd{3.5}{4B} & \hd{3.5}{9B} \\
\midrule
\textcolor{gray}{Teacher} & \textcolor{gray}{60.5} & \textcolor{gray}{60.5} & \textcolor{gray}{69.6} & \textcolor{gray}{69.6} & \textcolor{gray}{13.6} & \textcolor{gray}{13.6} & \textcolor{gray}{11.9} & \textcolor{gray}{11.9} & \textcolor{gray}{63.5} & \textcolor{gray}{63.5} & \textcolor{gray}{72.2} & \textcolor{gray}{72.2} \\
\midrule
Base & 50.8 & 54.3 & 61.9 & 66.2 & 13.1 & 11.5 & 13.4 & \second{11.3} & 51.2 & 55.1 & 64.7 & 68.6 \\
GRPO & 51.1 & 54.8 & 61.9 & 66.4 & 12.9 & 11.0 & 12.6 & 11.7 & 51.5 & 54.5 & 66.8 & 69.5 \\
OPD & 52.5 & 55.9 & 61.0 & 65.5 & \second{12.6} & \second{10.6} & 14.1 & 11.9 & 55.1 & 56.3 & 63.2 & 68.9 \\
S-OPD & 51.2 & 54.6 & 61.9 & 65.9 & 13.7 & 12.2 & 13.0 & 11.5 & 53.6 & 53.9 & 66.2 & 69.5 \\
OPD-Aha & \second{54.9} & \second{57.8} & \second{65.3} & \second{68.9} & 12.9 & 12.3 & \second{12.5} & 11.8 & \second{58.1} & \second{59.3} & \second{68.3} & \second{70.1} \\
\oursrow \method & \best{55.9} & \best{59.1} & \best{66.6} & \best{69.7} & \best{12.0} & \best{10.3} & \best{11.4} & \best{10.0} & \best{59.9} & \best{61.1} & \best{69.5} & \best{72.2} \\
\bottomrule
\end{tabular}

%% file: tables/ablation_main.tex
\begin{tabular}{@{}l c c c c c@{}}
\toprule
 & & \multicolumn{3}{c}{RealForm-Syn} & \hd{Gray-image}{answer rate} \\
\cmidrule(lr){3-5}
Setting & Avg. & Clean & \hd{Real-}{world} & Drop & (\%) \\
\midrule
\oursrow \method & 68.6 & 67.9 & 55.9 & 12.0 & 0.9 \\
\grp{6}{Without one part of \method} \\
\quad w/o evidence target & \cellcolor{red!14}66.5 & \cellcolor{red!14}64.5 & \cellcolor{red!14}52.7 & 11.8 & 3.8 \\
\quad w/o views & 68.2 & 67.9 & \cellcolor{red!14}54.9 & \cellcolor{red!14}12.9 & 1.0 \\
\grp{6}{Replacing excess selection in \method} \\
\quad random views & 68.0 & 67.0 & 55.2 & 11.8 & 2.3 \\
\midrule
OPD & \cellcolor{red!14}66.4 & \cellcolor{red!14}65.1 & \cellcolor{red!14}52.5 & \cellcolor{red!14}12.6 & 0.8 \\
\bottomrule
\end{tabular}

%% file: sec/6_conclusion.tex
\section{Conclusion}
\label{sec:conclusion}

Open VLMs lose accuracy on real-world forms of clean figures, a gap that RealForm measures across composited media and
real photos and that standard OPD leaves nearly unchanged. \method trains the student on harder views that stay clear
to the teacher, and reweights the clean-image target toward visual tokens still supported by each view. It thus
distills across forms without teaching unsupported answers, achieving the best results for all four students on
standard benchmarks, high-resolution perception and RealForm.

%% file: main.bbl
\begin{thebibliography}{80}
\providecommand{\natexlab}[1]{#1}
\providecommand{\url}[1]{\texttt{#1}}
\expandafter\ifx\csname urlstyle\endcsname\relax
  \providecommand{\doi}[1]{doi: #1}\else
  \providecommand{\doi}{doi: \begingroup \urlstyle{rm}\Url}\fi

\bibitem[Agarwal et~al.(2024)Agarwal, Vieillard, Zhou, Stanczyk, Ramos, Geist,
  and Bachem]{agarwal2024gkd}
Rishabh Agarwal, Nino Vieillard, Yongchao Zhou, Piotr Stanczyk, Sabela Ramos,
  Matthieu Geist, and Olivier Bachem.
\newblock On-policy distillation of language models: Learning from
  self-generated mistakes.
\newblock In \emph{International Conference on Learning Representations
  (ICLR)}, 2024.

\bibitem[Aniri et~al.(2026)Aniri, Bi, Liao, Jin, Tresp, Shen, Ma, and
  Chua]{opdv2026}
Aniri, Jinhe Bi, Peng Liao, Zengjie Jin, Volker Tresp, Fei Shen, Yunpu Ma, and
  Tat-Seng Chua.
\newblock {OPD-V}: Visual on-policy self-distillation with modality balance.
\newblock \emph{arXiv preprint arXiv:2608.05131}, 2026.

\bibitem[Bai et~al.(2025)Bai, Cai, Chen, Chen, Chen, Cheng, Deng, Ding,
  et~al.]{qwen3vl2025}
Shuai Bai, Yuxuan Cai, Ruizhe Chen, Keqin Chen, Xionghui Chen, Zesen Cheng,
  Lianghao Deng, Wei Ding, et~al.
\newblock {Qwen3-VL} technical report.
\newblock \emph{arXiv preprint arXiv:2511.21631}, 2025.

\bibitem[{Baidu ERNIE Team}(2025)]{ernie45}
{Baidu ERNIE Team}.
\newblock {ERNIE-4.5-VL-28B-A3B} model card.
\newblock \url{https://huggingface.co/baidu/ERNIE-4.5-VL-28B-A3B-PT}, 2025.

\bibitem[Barbu et~al.(2019)Barbu, Mayo, Alverio, Luo, Wang, Gutfreund,
  Tenenbaum, and Katz]{barbu2019objectnet}
Andrei Barbu, David Mayo, Julian Alverio, William Luo, Christopher Wang, Dan
  Gutfreund, Josh Tenenbaum, and Boris Katz.
\newblock {ObjectNet}: A large-scale bias-controlled dataset for pushing the
  limits of object recognition models.
\newblock In \emph{Advances in Neural Information Processing Systems
  (NeurIPS)}, 2019.

\bibitem[{Black Forest Labs}(2026)]{flux2klein}
{Black Forest Labs}.
\newblock {FLUX.2 [klein]}, 2026.
\newblock Image generation and editing models, released January 2026.

\bibitem[Chen et~al.(2019)Chen, Zhou, Koltun, and
  Kr{\"a}henb{\"u}hl]{chen2019lbc}
Dian Chen, Brady Zhou, Vladlen Koltun, and Philipp Kr{\"a}henb{\"u}hl.
\newblock Learning by cheating.
\newblock In \emph{Conference on Robot Learning (CoRL)}, 2019.

\bibitem[Chen et~al.(2024{\natexlab{a}})Chen, Li, Dong, Zhang, Zang, Chen,
  Duan, Wang, et~al.]{chen2024mmstar}
Lin Chen, Jinsong Li, Xiaoyi Dong, Pan Zhang, Yuhang Zang, Zehui Chen, Haodong
  Duan, Jiaqi Wang, et~al.
\newblock Are we on the right way for evaluating large vision-language models?
\newblock In \emph{Advances in Neural Information Processing Systems
  (NeurIPS)}, 2024{\natexlab{a}}.

\bibitem[Chen et~al.(2024{\natexlab{b}})Chen, Zhang, Peng, Zheng, Liu, Torr,
  and Stiefelhagen]{chen2024rodla}
Yufan Chen, Jiaming Zhang, Kunyu Peng, Junwei Zheng, Ruiping Liu, Philip Torr,
  and Rainer Stiefelhagen.
\newblock {RoDLA}: Benchmarking the robustness of document layout analysis
  models.
\newblock In \emph{IEEE/CVF Conference on Computer Vision and Pattern
  Recognition (CVPR)}, pages 15556--15566, 2024{\natexlab{b}}.

\bibitem[Du et~al.(2025)Du, Chen, Ying, and Chen]{docptbench2025}
Yongkun Du, Pinxuan Chen, Xuye Ying, and Zhineng Chen.
\newblock {DocPTBench}: Benchmarking end-to-end photographed document parsing
  and translation.
\newblock \emph{arXiv preprint arXiv:2511.18434}, 2025.

\bibitem[Favero et~al.(2024)Favero, Zancato, Trager, Choudhary, Perera,
  Achille, Swaminathan, and Soatto]{favero2024m3id}
Alessandro Favero, Luca Zancato, Matthew Trager, Siddharth Choudhary,
  Pramuditha Perera, Alessandro Achille, Ashwin Swaminathan, and Stefano
  Soatto.
\newblock Multi-modal hallucination control by visual information grounding.
\newblock In \emph{IEEE/CVF Conference on Computer Vision and Pattern
  Recognition (CVPR)}, pages 14303--14312, 2024.

\bibitem[Feng(2026)]{vcopd2026}
Jiahe Feng.
\newblock {VC-OPD}: Visual counterfactual on-policy distillation for grounded
  vision-language reasoning.
\newblock In \emph{Advances in Neural Information Processing Systems
  (NeurIPS)}, 2026.
\newblock arXiv:2610.14939.

\bibitem[Feng et~al.(2025)Feng, Wang, Zhang, Guo, Zhou, Chen, Li, and
  Yin]{mathreal2025}
Jun Feng, Zixin Wang, Zhentao Zhang, Yue Guo, Zhihan Zhou, Xiuyi Chen, Zhenyang
  Li, and Dawei Yin.
\newblock {MathReal}: We keep it real! a real scene benchmark for evaluating
  math reasoning in multimodal large language models.
\newblock \emph{arXiv preprint arXiv:2508.06009}, 2025.

\bibitem[Fu et~al.(2024)Fu, Hu, Li, Feng, Wang, Lin, Roth, Smith, Ma, and
  Krishna]{fu2024blink}
Xingyu Fu, Yushi Hu, Bangzheng Li, Yu Feng, Haoyu Wang, Xudong Lin, Dan Roth,
  Noah~A. Smith, Wei-Chiu Ma, and Ranjay Krishna.
\newblock {BLINK}: Multimodal large language models can see but not perceive.
\newblock In \emph{European Conference on Computer Vision (ECCV)}, 2024.

\bibitem[{Gemma Team}(2026)]{gemma4}
{Gemma Team}.
\newblock Gemma 4 model card.
\newblock \url{https://ai.google.dev/gemma/docs/core/model_card_4}, 2026.

\bibitem[{GLM-V Team}(2025)]{glm45v}
{GLM-V Team}.
\newblock {GLM-4.5V} and {GLM-4.1V-Thinking}: Towards versatile multimodal
  reasoning with scalable reinforcement learning.
\newblock \emph{arXiv preprint arXiv:2507.01006}, 2025.

\bibitem[Gu et~al.(2024)Gu, Dong, Wei, and Huang]{gu2024minillm}
Yuxian Gu, Li Dong, Furu Wei, and Minlie Huang.
\newblock {MiniLLM}: Knowledge distillation of large language models.
\newblock In \emph{International Conference on Learning Representations
  (ICLR)}, 2024.

\bibitem[Hendrycks and Dietterich(2019)]{hendrycks2019imagenetc}
Dan Hendrycks and Thomas Dietterich.
\newblock Benchmarking neural network robustness to common corruptions and
  perturbations.
\newblock In \emph{International Conference on Learning Representations
  (ICLR)}, 2019.

\bibitem[Hendrycks et~al.(2020)Hendrycks, Mu, Cubuk, Zoph, Gilmer, and
  Lakshminarayanan]{hendrycks2020augmix}
Dan Hendrycks, Norman Mu, Ekin~Dogus Cubuk, Barret Zoph, Justin Gilmer, and
  Balaji Lakshminarayanan.
\newblock {AugMix}: A simple data processing method to improve robustness and
  uncertainty.
\newblock In \emph{International Conference on Learning Representations
  (ICLR)}, 2020.

\bibitem[Hendrycks et~al.(2021)Hendrycks, Basart, Mu, Kadavath, Wang, Dorundo,
  Desai, Zhu, Parajuli, Guo, Song, Steinhardt, and
  Gilmer]{hendrycks2021imagenetr}
Dan Hendrycks, Steven Basart, Norman Mu, Saurav Kadavath, Frank Wang, Evan
  Dorundo, Rahul Desai, Tyler Zhu, Samyak Parajuli, Mike Guo, Dawn Song, Jacob
  Steinhardt, and Justin Gilmer.
\newblock The many faces of robustness: A critical analysis of
  out-of-distribution generalization.
\newblock In \emph{IEEE/CVF International Conference on Computer Vision
  (ICCV)}, pages 8340--8349, 2021.

\bibitem[Huang et~al.(2026)Huang, Fang, Xiao, and Du]{pvd2026}
Shanfeng Huang, Zhou Fang, Song Xiao, and Hai Du.
\newblock Progressive-view on-policy distillation for regional-to-global
  transfer in multimodal {LLMs}.
\newblock \emph{arXiv preprint arXiv:2609.32333}, 2026.

\bibitem[Jeddi et~al.(2026)Jeddi, Zhang, Gerigk, Karaimer, Azadani, Luo, Le,
  Aminian, et~al.]{scopd2026}
Ahmadreza Jeddi, Enming Zhang, Jasper Gerigk, Hakki Karaimer, Mozhgan~Nasr
  Azadani, Jiayun Luo, Minh~Ngoc Le, Gholamali Aminian, et~al.
\newblock {SCOPD}: Sparse-context on-policy self-distillation for efficient
  vision-language models.
\newblock \emph{arXiv preprint arXiv:2609.34044}, 2026.

\bibitem[Kim and Rush(2016)]{kim2016seqkd}
Yoon Kim and Alexander~M. Rush.
\newblock Sequence-level knowledge distillation.
\newblock In \emph{Conference on Empirical Methods in Natural Language
  Processing (EMNLP)}, 2016.

\bibitem[{Kimi Team}(2025)]{kimivl}
{Kimi Team}.
\newblock {Kimi-VL} technical report.
\newblock \emph{arXiv preprint arXiv:2504.07491}, 2025.

\bibitem[Ko et~al.(2024)Ko, Kim, Chen, and Yun]{ko2024distillm}
Jongwoo Ko, Sungnyun Kim, Tianyi Chen, and Se-Young Yun.
\newblock {DistiLLM}: Towards streamlined distillation for large language
  models.
\newblock In \emph{International Conference on Machine Learning (ICML)}, 2024.

\bibitem[Kurakin et~al.(2017)Kurakin, Goodfellow, and
  Bengio]{kurakin2017physical}
Alexey Kurakin, Ian Goodfellow, and Samy Bengio.
\newblock Adversarial examples in the physical world.
\newblock In \emph{International Conference on Learning Representations (ICLR)
  Workshop}, 2017.

\bibitem[Leng et~al.(2024)Leng, Zhang, Chen, Li, Lu, Miao, and
  Bing]{leng2024vcd}
Sicong Leng, Hang Zhang, Guanzheng Chen, Xin Li, Shijian Lu, Chunyan Miao, and
  Lidong Bing.
\newblock Mitigating object hallucinations in large vision-language models
  through visual contrastive decoding.
\newblock In \emph{IEEE/CVF Conference on Computer Vision and Pattern
  Recognition (CVPR)}, pages 13872--13882, 2024.

\bibitem[Li et~al.(2024)Li, Zhang, Zhang, Wu, Tian, Sun, Lu, Liu,
  et~al.]{rbench2024}
Chunyi Li, Jianbo Zhang, Zicheng Zhang, Haoning Wu, Yuan Tian, Wei Sun, Guo Lu,
  Xiaohong Liu, et~al.
\newblock {R-Bench}: Are your large multimodal model robust to real-world
  corruptions?
\newblock \emph{arXiv preprint arXiv:2410.05474}, 2024.

\bibitem[Li et~al.(2026{\natexlab{a}})Li, Yang, Zhang, Xu, Dong, and
  Zhang]{ltopd2026}
Junxian Li, Ruixuan Yang, Tianao Zhang, Tiange Xu, Weisheng Dong, and Yulun
  Zhang.
\newblock Fewer tokens, more self-teaching: On-policy self-distillation for
  extreme visual token reduction.
\newblock \emph{arXiv preprint arXiv:2609.32353}, 2026{\natexlab{a}}.

\bibitem[Li et~al.(2026{\natexlab{b}})Li, Liang, Tian, Wang, Zhang, Yin, Fu,
  Torr, and Vasconcelos]{s2vopd2026}
Yijiang Li, Yijun Liang, Yunjie Tian, Bingyang Wang, Ke Zhang, Zhenfei Yin, Di
  Fu, Philip Torr, and Nuno Vasconcelos.
\newblock Self-supervised visual on-policy distillation.
\newblock \emph{arXiv preprint arXiv:2608.14144}, 2026{\natexlab{b}}.

\bibitem[Liang et~al.(2024)Liang, Sun, Zhou, Chen, Qiang, Lin, Li, Yang, Zhou,
  and Zhang]{mathscape2024}
Hao Liang, Linzhuang Sun, Minxuan Zhou, Zirong Chen, Meiyi Qiang, Mingan Lin,
  Tianpeng Li, Fan Yang, Zenan Zhou, and Wentao Zhang.
\newblock {MathScape}: Benchmarking multimodal large language models in
  real-world mathematical contexts.
\newblock \emph{arXiv preprint arXiv:2408.07543}, 2024.

\bibitem[Liang et~al.(2026)Liang, Tian, Li, Jia, Wang, Zhou, and Fu]{vcsd2026}
Yijun Liang, Yunjie Tian, Yijiang Li, Yuqi Jia, Kaishen Wang, Tianyi Zhou, and
  Di Fu.
\newblock Visual contrastive self-distillation.
\newblock \emph{arXiv preprint arXiv:2607.21556}, 2026.

\bibitem[Liu et~al.(2026{\natexlab{a}})Liu, Lv, Li, Zhu, Wang, Zhang, Chen, Li,
  et~al.]{vaopd2026}
Ruiqi Liu, Xiaolei Lv, Gengsheng Li, Ximo Zhu, Zhiheng Wang, Zhengbo Zhang,
  Junkai Chen, Zhiheng Li, et~al.
\newblock Visual-advantage on-policy distillation for vision-language models.
\newblock \emph{arXiv preprint arXiv:2605.21924}, 2026{\natexlab{a}}.

\bibitem[Liu et~al.(2026{\natexlab{b}})Liu, Tian, Duan, He, Yang, Zhang, and
  Shi]{sopd2026}
Siyuan Liu, Kanghui Tian, Yue Duan, Yutao He, Shangdong Yang, Jian Zhang, and
  Yinghuan Shi.
\newblock Is better teacher supervision enough? unlocking student-side learning
  in multimodal on-policy distillation.
\newblock \emph{arXiv preprint arXiv:2609.39120}, 2026{\natexlab{b}}.

\bibitem[Lopez-Paz et~al.(2016)Lopez-Paz, Bottou, Sch{\"o}lkopf, and
  Vapnik]{lopezpaz2016unifying}
David Lopez-Paz, L{\'e}on Bottou, Bernhard Sch{\"o}lkopf, and Vladimir Vapnik.
\newblock Unifying distillation and privileged information.
\newblock In \emph{International Conference on Learning Representations
  (ICLR)}, 2016.

\bibitem[Lu and {Thinking Machines Lab}(2025)]{tm2025opd}
Kevin Lu and {Thinking Machines Lab}.
\newblock On-policy distillation.
\newblock Thinking Machines Lab: Connectionism,
  \url{https://thinkingmachines.ai/blog/on-policy-distillation/}, 2025.

\bibitem[Lu et~al.(2021)Lu, Gong, Jiang, Qiu, Huang, Liang, and
  Zhu]{lu2021intergps}
Pan Lu, Ran Gong, Shibiao Jiang, Liang Qiu, Siyuan Huang, Xiaodan Liang, and
  Song-Chun Zhu.
\newblock {Inter-GPS}: Interpretable geometry problem solving with formal
  language and symbolic reasoning.
\newblock In \emph{Annual Meeting of the Association for Computational
  Linguistics (ACL)}, 2021.

\bibitem[Lu et~al.(2024)Lu, Bansal, Xia, Liu, Li, Hajishirzi, Cheng, Chang,
  Galley, and Gao]{lu2024mathvista}
Pan Lu, Hritik Bansal, Tony Xia, Jiacheng Liu, Chunyuan Li, Hannaneh
  Hajishirzi, Hao Cheng, Kai-Wei Chang, Michel Galley, and Jianfeng Gao.
\newblock {MathVista}: Evaluating mathematical reasoning of foundation models
  in visual contexts.
\newblock In \emph{International Conference on Learning Representations
  (ICLR)}, 2024.

\bibitem[Ma et~al.(2018)Ma, Shu, Bai, Wang, and Samaras]{ma2018docunet}
Ke Ma, Zhixin Shu, Xue Bai, Jue Wang, and Dimitris Samaras.
\newblock {DocUNet}: Document image unwarping via a stacked {U-Net}.
\newblock In \emph{IEEE/CVF Conference on Computer Vision and Pattern
  Recognition (CVPR)}, pages 4700--4709, 2018.

\bibitem[{MathLLMs}(2025)]{mathvisionwild}
{MathLLMs}.
\newblock {MathVision-Wild}.
\newblock \url{https://huggingface.co/datasets/MathLLMs/MathVision-Wild}, 2025.

\bibitem[{Meta AI}(2025)]{llama4}
{Meta AI}.
\newblock {Llama-4-Scout-17B-16E-Instruct} model card.
\newblock
  \url{https://huggingface.co/meta-llama/Llama-4-Scout-17B-16E-Instruct}, 2025.

\bibitem[{Mistral AI}(2025)]{mistralsmall32}
{Mistral AI}.
\newblock {Mistral-Small-3.2-24B-Instruct-2506} model card.
\newblock
  \url{https://huggingface.co/mistralai/Mistral-Small-3.2-24B-Instruct-2506},
  2025.

\bibitem[Moured et~al.(2025)Moured, Chen, Liu, Rei{\ss}, Torr, Zhang, and
  Stiefelhagen]{moured2025chaos}
Omar Moured, Yufan Chen, Ruiping Liu, Simon Rei{\ss}, Philip Torr, Jiaming
  Zhang, and Rainer Stiefelhagen.
\newblock {CHAOS}: Chart analysis with outlier samples.
\newblock \emph{arXiv preprint arXiv:2505.17235}, 2025.

\bibitem[Qiao et~al.(2024)Qiao, Tan, Dong, Wu, Sun, Song, GongQue, Lei,
  et~al.]{qiao2024wemath}
Runqi Qiao, Qiuna Tan, Guanting Dong, Minhui Wu, Chong Sun, Xiaoshuai Song,
  Zhuoma GongQue, Shanglin Lei, et~al.
\newblock {We-Math}: Does your large multimodal model achieve human-like
  mathematical reasoning?
\newblock \emph{arXiv preprint arXiv:2407.01284}, 2024.

\bibitem[Qiu et~al.(2026)Qiu, Li, Zhang, Gong, and Tan]{opdaha2026}
Chenhao Qiu, Dawei Li, Yechao Zhang, Lei Gong, and Zhen Tan.
\newblock {OPD-Aha}: From linguistic momentum to visual reflection in
  multimodal on-policy distillation.
\newblock \emph{arXiv preprint arXiv:2609.16459}, 2026.

\bibitem[Qiu et~al.(2024)Qiu, Zhu, Shi, Wenzel, Tang, Zhao, Li, and
  Li]{qiu2024mmshift}
Jielin Qiu, Yi Zhu, Xingjian Shi, Florian Wenzel, Zhiqiang Tang, Ding Zhao, Bo
  Li, and Mu Li.
\newblock Benchmarking robustness of multimodal image-text models under
  distribution shift.
\newblock \emph{Journal of Data-centric Machine Learning Research (DMLR)},
  2024.

\bibitem[{Qwen Team}(2024)]{qwen25report}
{Qwen Team}.
\newblock {Qwen2.5} technical report.
\newblock \emph{arXiv preprint arXiv:2412.15115}, 2024.

\bibitem[{Qwen Team}(2026)]{qwen35}
{Qwen Team}.
\newblock {Qwen3.5} model release.
\newblock \url{https://huggingface.co/Qwen/Qwen3.5-27B}, 2026.

\bibitem[Schulman(2020)]{schulman2020klapprox}
John Schulman.
\newblock Approximating {KL} divergence.
\newblock \url{http://joschu.net/blog/kl-approx.html}, 2020.

\bibitem[Shao et~al.(2024)Shao, Wang, Zhu, Xu, Song, Bi, Zhang, Zhang, Li, Wu,
  and Guo]{shao2024deepseekmath}
Zhihong Shao, Peiyi Wang, Qihao Zhu, Runxin Xu, Junxiao Song, Xiao Bi, Haowei
  Zhang, Mingchuan Zhang, Y.~K. Li, Y. Wu, and Daya Guo.
\newblock {DeepSeekMath}: Pushing the limits of mathematical reasoning in open
  language models.
\newblock \emph{arXiv preprint arXiv:2402.03300}, 2024.

\bibitem[Song and Zheng(2026)]{opdsurvey2026}
Mingyang Song and Mao Zheng.
\newblock A survey of on-policy distillation for large language models.
\newblock \emph{arXiv preprint arXiv:2604.00626}, 2026.

\bibitem[Tang et~al.(2026)Tang, Chen, Wei, Xu, Liu, Wu, Xie, Wu, Zhang, and
  Chen]{robustr12025}
Jiaqi Tang, Jianmin Chen, Wei Wei, Xiaogang Xu, Runtao Liu, Xiangyu Wu, Qipeng
  Xie, Jiafei Wu, Lei Zhang, and Qifeng Chen.
\newblock {Robust-R1}: Degradation-aware reasoning for robust visual
  understanding.
\newblock In \emph{AAAI Conference on Artificial Intelligence (AAAI)}, 2026.

\bibitem[Wang et~al.(2025{\natexlab{a}})Wang, Tang, Liao, Feng, Liu, Fei, Lu,
  Wang, et~al.]{wilddoc2025}
An-Lan Wang, Jingqun Tang, Lei Liao, Hao Feng, Qi Liu, Xiang Fei, Jinghui Lu,
  Han Wang, et~al.
\newblock {WildDoc}: How far are we from achieving comprehensive and robust
  document understanding in the wild?
\newblock In \emph{Conference on Empirical Methods in Natural Language
  Processing (EMNLP)}, 2025{\natexlab{a}}.

\bibitem[Wang et~al.(2019)Wang, Ge, Lipton, and Xing]{wang2019imagenetsketch}
Haohan Wang, Songwei Ge, Zachary Lipton, and Eric~P. Xing.
\newblock Learning robust global representations by penalizing local predictive
  power.
\newblock In \emph{Advances in Neural Information Processing Systems
  (NeurIPS)}, 2019.

\bibitem[Wang et~al.(2025{\natexlab{b}})Wang, Qu, Huang, Chu, Lin, and
  Chen]{vlrethinker2025}
Haozhe Wang, Chao Qu, Zuming Huang, Wei Chu, Fangzhen Lin, and Wenhu Chen.
\newblock {VL-Rethinker}: Incentivizing self-reflection of vision-language
  models with reinforcement learning.
\newblock \emph{arXiv preprint arXiv:2504.08837}, 2025{\natexlab{b}}.

\bibitem[Wang et~al.(2024)Wang, Pan, Shi, Lu, Ren, Zhou, Zhan, and
  Li]{wang2024mathvision}
Ke Wang, Junting Pan, Weikang Shi, Zimu Lu, Houxing Ren, Aojun Zhou, Mingjie
  Zhan, and Hongsheng Li.
\newblock Measuring multimodal mathematical reasoning with {MATH-Vision}
  dataset.
\newblock In \emph{Advances in Neural Information Processing Systems (NeurIPS),
  Datasets and Benchmarks Track}, 2024.

\bibitem[Wang et~al.(2026{\natexlab{a}})Wang, Zhang, Tang, Cheng, and
  Wei]{nopd2026}
Shuai Wang, Daoan Zhang, Zhe Tang, Hao Cheng, and Jiaheng Wei.
\newblock Self-boosting vision-language models with {Noisy Student} on-policy
  self-distillation.
\newblock \emph{arXiv preprint arXiv:2607.23125}, 2026{\natexlab{a}}.

\bibitem[Wang et~al.(2025{\natexlab{c}})Wang, Ding, Zeng, Zhou, Shen, Luo, Yu,
  and Tao]{wang2025hrbench}
Wenbin Wang, Liang Ding, Minyan Zeng, Xiabin Zhou, Li Shen, Yong Luo, Wei Yu,
  and Dacheng Tao.
\newblock Divide, conquer and combine: A training-free framework for
  high-resolution image perception in multimodal large language models.
\newblock In \emph{AAAI}, 2025{\natexlab{c}}.

\bibitem[Wang et~al.(2025{\natexlab{d}})]{internvl35}
Weiyun Wang et~al.
\newblock {InternVL3.5}: Advancing open-source multimodal models in
  versatility, reasoning, and efficiency.
\newblock \emph{arXiv preprint arXiv:2508.18265}, 2025{\natexlab{d}}.

\bibitem[Wang et~al.(2026{\natexlab{b}})Wang, Cui, Zhao, Yang, Zhu, and
  Shao]{mathsight2025}
Yuandong Wang, Yao Cui, Yuxin Zhao, Zhen Yang, Yangfu Zhu, and Zhenzhou Shao.
\newblock {MathSight}: A benchmark exploring have vision-language models really
  seen in university-level mathematical reasoning?
\newblock In \emph{Annual Meeting of the Association for Computational
  Linguistics (ACL)}, 2026{\natexlab{b}}.

\bibitem[Wei et~al.(2026)Wei, He, Lan, Dong, Cai, Li, Zhu, Wang, Kong, Wang,
  Zhang, and Huang]{wei2026zoombench}
Lai Wei, Liangbo He, Jun Lan, Lingzhong Dong, Yutong Cai, Siyuan Li, Huijia
  Zhu, Weiqiang Wang, Linghe Kong, Yue Wang, Zhuosheng Zhang, and Weiran Huang.
\newblock Zooming without zooming: Region-to-image distillation for
  fine-grained multimodal perception.
\newblock \emph{arXiv preprint arXiv:2602.11858}, 2026.

\bibitem[Weihs et~al.(2021)Weihs, Jain, Liu, Salvador, Lazebnik, Kembhavi, and
  Schwing]{weihs2021advisor}
Luca Weihs, Unnat Jain, Iou-Jen Liu, Jordi Salvador, Svetlana Lazebnik,
  Aniruddha Kembhavi, and Alexander Schwing.
\newblock Bridging the imitation gap by adaptive insubordination.
\newblock In \emph{Advances in Neural Information Processing Systems
  (NeurIPS)}, 2021.

\bibitem[Wu et~al.(2025)Wu, Li, Zhou, Lin, Gao, Yan, ming Yin, Bai,
  et~al.]{qwenimage2025}
Chenfei Wu, Jiahao Li, Jingren Zhou, Junyang Lin, Kaiyuan Gao, Kun Yan, Sheng
  ming Yin, Shuai Bai, et~al.
\newblock {Qwen-Image} technical report.
\newblock \emph{arXiv preprint arXiv:2508.02324}, 2025.

\bibitem[Wu et~al.(2024)Wu, Zhang, Zhang, Chen, Liao, Wang, Li, Sun, Yan, Zhai,
  and Lin]{wu2024qbench}
Haoning Wu, Zicheng Zhang, Erli Zhang, Chaofeng Chen, Liang Liao, Annan Wang,
  Chunyi Li, Wenxiu Sun, Qiong Yan, Guangtao Zhai, and Weisi Lin.
\newblock {Q-Bench}: A benchmark for general-purpose foundation models on
  low-level vision.
\newblock In \emph{International Conference on Learning Representations
  (ICLR)}, 2024.

\bibitem[Wu and Xie(2024)]{wu2024vstar}
Penghao Wu and Saining Xie.
\newblock {V*}: Guided visual search as a core mechanism in multimodal {LLMs}.
\newblock In \emph{CVPR}, 2024.

\bibitem[Xiao et~al.(2024)Xiao, Sun, Liu, and Wang]{xiao2024logicvista}
Yijia Xiao, Edward Sun, Tianyu Liu, and Wei Wang.
\newblock {LogicVista}: Multimodal {LLM} logical reasoning benchmark in visual
  contexts.
\newblock \emph{arXiv preprint arXiv:2407.04973}, 2024.

\bibitem[Xie et~al.(2020)Xie, Luong, Hovy, and Le]{xie2020noisystudent}
Qizhe Xie, Minh-Thang Luong, Eduard Hovy, and Quoc~V. Le.
\newblock Self-training with {Noisy Student} improves {ImageNet}
  classification.
\newblock In \emph{IEEE/CVF Conference on Computer Vision and Pattern
  Recognition (CVPR)}, 2020.

\bibitem[Xiong et~al.(2026)Xiong, Xue, and Lin]{pcd2026}
Feng Xiong, Leyan Xue, and Hongyu Lin.
\newblock Correcting what you cannot see: Credit assignment for perception
  distillation in multimodal reasoners.
\newblock \emph{arXiv preprint arXiv:2607.28336}, 2026.

\bibitem[Xu et~al.(2026)Xu, Sang, Zhou, He, Wang, and Geramifard]{tip2026}
Yuanda Xu, Hejian Sang, Zhengze Zhou, Ran He, Zhipeng Wang, and Alborz
  Geramifard.
\newblock {TIP}: Token importance in on-policy distillation.
\newblock \emph{arXiv preprint arXiv:2604.14084}, 2026.

\bibitem[Xue et~al.(2026)Xue, Xiong, Ma, and Zhang]{fpopd2026}
Leyan Xue, Feng Xiong, Mingjun Ma, and Changqing Zhang.
\newblock Distill what the student can see: {Fisher}-projected on-policy
  distillation for vision-language models.
\newblock \emph{arXiv preprint arXiv:2608.01263}, 2026.

\bibitem[Yu et~al.(2026)Yu, Li, Si, Zhang, Xu, Wang, Dong, Tuo, Zeng, Feng,
  Wang, Shi, Hu, Yue, Wang, and Yan]{dopd2026}
Xinlei Yu, Gen Li, Qingyi Si, Guibin Zhang, Yuqi Xu, Congcong Wang, Shuai Dong,
  Kaiwen Tuo, Xiangyu Zeng, Kaituo Feng, Qunzhong Wang, Yang Shi, Xiaobin Hu,
  Xiangyu Yue, Jiaqi Wang, and Shuicheng Yan.
\newblock {DOPD}: Dual on-policy distillation.
\newblock \emph{arXiv preprint arXiv:2606.30626}, 2026.

\bibitem[Yuan et~al.(2026)Yuan, Lou, Yu, Lin, Sun, Han, and Lu]{visionopd2026}
Qianhao Yuan, Jie Lou, Xing Yu, Hongyu Lin, Le Sun, Xianpei Han, and Yaojie Lu.
\newblock {Vision-OPD}: Learning to see fine details for multimodal {LLMs} via
  on-policy self-distillation.
\newblock \emph{arXiv preprint arXiv:2605.18740}, 2026.

\bibitem[Yue et~al.(2024)Yue, Ni, Zhang, Zheng, Liu, Zhang, Stevens, Jiang,
  et~al.]{yue2024mmmu}
Xiang Yue, Yuansheng Ni, Kai Zhang, Tianyu Zheng, Ruoqi Liu, Ge Zhang, Samuel
  Stevens, Dongfu Jiang, et~al.
\newblock {MMMU}: A massive multi-discipline multimodal understanding and
  reasoning benchmark for expert {AGI}.
\newblock In \emph{IEEE/CVF Conference on Computer Vision and Pattern
  Recognition (CVPR)}, 2024.

\bibitem[Yue et~al.(2025)Yue, Zheng, Ni, Wang, Zhang, Tong, Sun, Yu,
  et~al.]{yue2024mmmupro}
Xiang Yue, Tianyu Zheng, Yuansheng Ni, Yubo Wang, Kai Zhang, Shengbang Tong,
  Yuxuan Sun, Botao Yu, et~al.
\newblock {MMMU-Pro}: A more robust multi-discipline multimodal understanding
  benchmark.
\newblock In \emph{Annual Meeting of the Association for Computational
  Linguistics (ACL)}, 2025.

\bibitem[Zhang et~al.(2024{\natexlab{a}})Zhang, Pang, Du, Ren, Li, and
  Lin]{mmcbench2024}
Jiawei Zhang, Tianyu Pang, Chao Du, Yi Ren, Bo Li, and Min Lin.
\newblock Benchmarking large multimodal models against common corruptions.
\newblock \emph{arXiv preprint arXiv:2401.11943}, 2024{\natexlab{a}}.

\bibitem[Zhang et~al.(2026)Zhang, Li, Shao, Li, Lu, Yao, Lin, Jiao, Lu, Liu,
  Zhang, and Yu]{vad2026}
Kangning Zhang, Yixing Li, Shuai Shao, Qingyao Li, Zhengxi Lu, Zhiyuan Yao,
  Jianghao Lin, Wenxiang Jiao, Yuan Lu, Weiwen Liu, Weinan Zhang, and Yong Yu.
\newblock {VAD}: Attributing visual evidence for target reconstruction in
  multimodal on-policy distillation.
\newblock \emph{arXiv preprint arXiv:2607.28590}, 2026.

\bibitem[Zhang et~al.(2024{\natexlab{b}})Zhang, Jiang, Zhang, Lin, Guo, Qiu,
  Zhou, Lu, et~al.]{zhang2024mathverse}
Renrui Zhang, Dongzhi Jiang, Yichi Zhang, Haokun Lin, Ziyu Guo, Pengshuo Qiu,
  Aojun Zhou, Pan Lu, et~al.
\newblock {MathVerse}: Does your multi-modal {LLM} truly see the diagrams in
  visual math problems?
\newblock In \emph{European Conference on Computer Vision (ECCV)},
  2024{\natexlab{b}}.

\bibitem[Zhang et~al.(2025)]{zhang2025mmerealworld}
Yi-Fan Zhang et~al.
\newblock {MME-RealWorld}: Could your multimodal {LLM} challenge
  high-resolution real-world scenarios that are difficult for humans?
\newblock In \emph{ICLR}, 2025.

\bibitem[Zhu et~al.(2026)Zhu, Wang, Wen, Zhang, Zhang, Liu, Chen, Wu, Yang, and
  Jiang]{rpopsd2026}
Qihui Zhu, Yuchen Wang, Zijian Wen, Tao Zhang, Mengjie Zhang, Yang Liu,
  Shuangwu Chen, Siying Wu, Jian Yang, and Xiaofeng Jiang.
\newblock {RP-OPSD}: Resolution-privileged on-policy self-distillation for
  multimodal large language models.
\newblock \emph{arXiv preprint arXiv:2607.24447}, 2026.

\bibitem[Zou et~al.(2025)Zou, Guo, Yang, Zhang, Hu, and Zhang]{zou2024dynamath}
Chengke Zou, Xingang Guo, Rui Yang, Junyu Zhang, Bin Hu, and Huan Zhang.
\newblock {DynaMath}: A dynamic visual benchmark for evaluating mathematical
  reasoning robustness of vision language models.
\newblock In \emph{International Conference on Learning Representations
  (ICLR)}, 2025.

\end{thebibliography}
